\documentclass[10pt,twocolumn,letterpaper]{article}

\usepackage{cvpr}
\usepackage{times}
\usepackage{epsfig}
\usepackage{graphicx}
\usepackage{amsmath}
\usepackage{amssymb}
\usepackage{multirow}
\usepackage{appendix}
\usepackage{subfigure}
\usepackage{algorithm,algpseudocode}
\usepackage{caption}
\newcommand{\cev}[1]{\reflectbox{\ensuremath{\vec{\reflectbox{\ensuremath{#1}}}}}}

\usepackage[pagebackref=true,breaklinks=true,letterpaper=true,colorlinks,bookmarks=false]{hyperref}

\cvprfinalcopy % *** Uncomment this line for the final submission

\ifcvprfinal\fi
\begin{document}

%%%%%%%%% TITLE
\title{Weakly Supervised Dense Video Captioning}
\author{Zhiqiang Shen$^{\dag\thanks{This work was done when Zhiqiang Shen was an intern at Intel Labs China. Jianguo Li and Yu-Gang Jiang are the corresponding authors.}}$, ~ Jianguo Li$^\ddag$, ~ Zhou Su$^\ddag$, ~ Minjun Li$^\dag$ \\ Yurong Chen$^\ddag$, ~ Yu-Gang Jiang$^\dag$, ~ Xiangyang Xue$^\dag$ \\
%	$^\dag$Shanghai Key Laboratory of Intelligent Information Processing,\\
%	School of Computer Science, Fudan University
     $^\dag$Shanghai Key Laboratory of Intelligent Information Processing \\
     School of Computer Science, Fudan University \\
	$^\ddag$Intel Labs China\\
	\tt\small $^\dag$\{zhiqiangshen13, minjunli13, ygj, xyxue\}@fudan.edu.cn \\ \tt\small $^\ddag$\{jianguo.li, zhou.su, yurong.chen\}@intel.com
}

\maketitle
\thispagestyle{empty}

%%%%%%%%% ABSTRACT
\begin{abstract}
This paper focuses on a novel and challenging vision task, dense video captioning, which aims to automatically describe a video clip with multiple informative and diverse caption sentences. The proposed method is trained without explicit annotation of fine-grained sentence to video region-sequence correspondence, but is only based on \emph{weak} video-level sentence annotations. It differs from existing video captioning systems in three technical aspects. First, we propose lexical fully convolutional neural networks (Lexical-FCN) with weakly supervised multi-instance multi-label learning to weakly link video regions with lexical labels. Second, we introduce a novel submodular maximization scheme to generate multiple informative and diverse region-sequences based on the Lexical-FCN outputs. A winner-takes-all scheme is adopted to weakly associate sentences to region-sequences in the training phase. Third, a sequence-to-sequence learning based language model is trained with the weakly supervised information obtained through the association process. We show that the proposed method can not only produce informative and diverse dense captions, but also outperform state-of-the-art single video captioning methods by a large margin.

\end{abstract}

%%%%%%%%% BODY TEXT
\section{Introduction}
Automatically describing images or videos with natural language sentences has recently received significant attention in the computer vision community.
For images, researchers have investigated image captioning with one sentence~\cite{xu2015show,vinyals2015show,donahue2015long,anne2016deep,fang2015captions, mao2014deep,venugopalan2016captioning} or multiple sentences~\cite{karpathy2015deep,johnson2015densecap,plummer2015flickr30k}.
For videos, most of the works focused on generating only one caption for a short video clip using methods based on mean pooling of features over frames~\cite{venugopalan2014translating}, the soft-attention scheme \cite{yao2015describing}, or visual-semantic embedding between visual feature and language~\cite{pan2015jointly}.
Some recent works further considered the video temporal structure, such as the sequence-to-sequence learning (S2VT)~\cite{venugopalan2015sequence} and hierarchical recurrent neural encoder \cite{pan2015hierarchical}.
\begin{figure}[]
	\centering
	\includegraphics[width=0.92\linewidth]{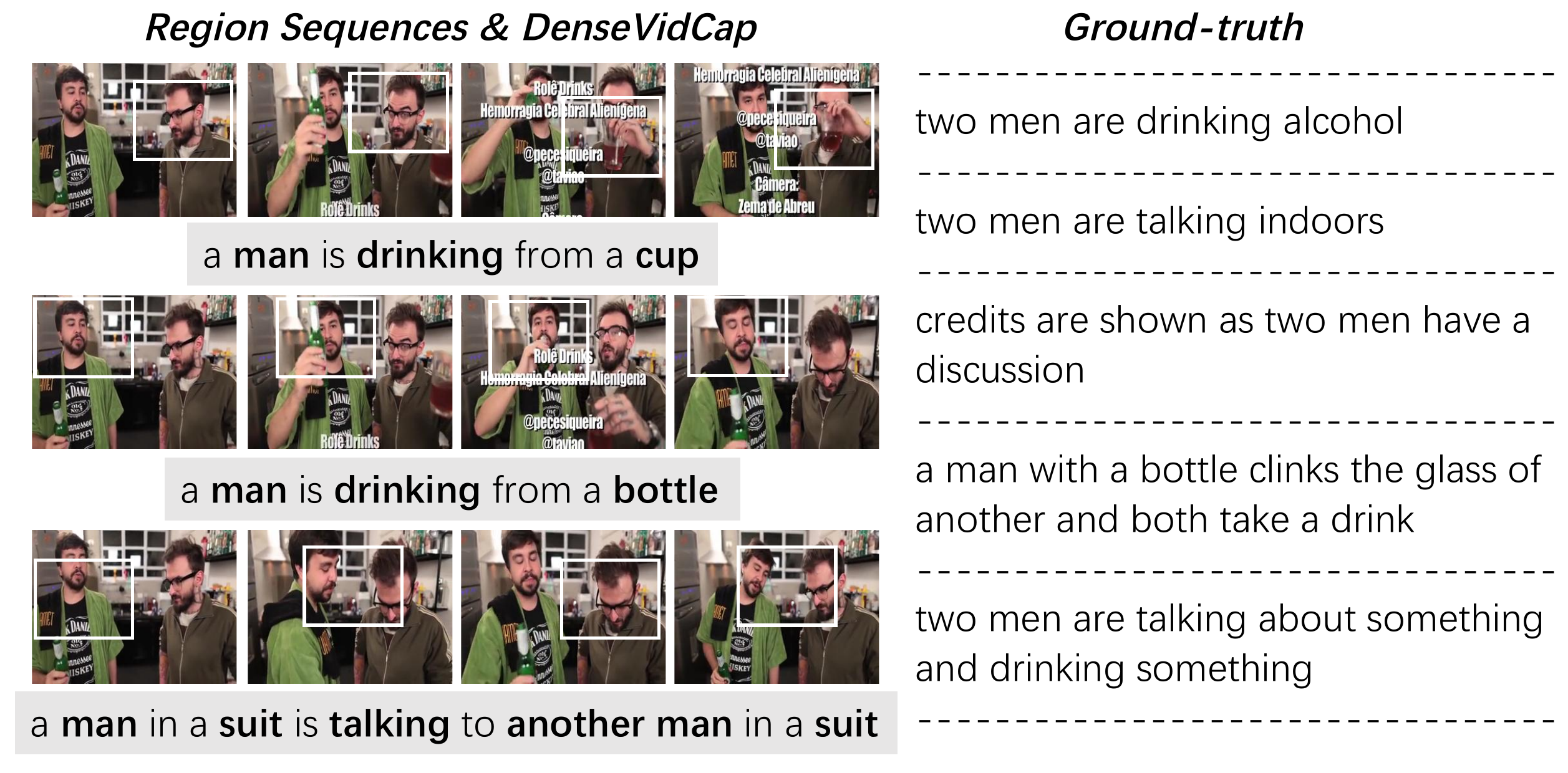}
	\vspace{-1.5 ex}
	\caption{Illustration of dense video captioning ({\em DenseVidCap}). Each region-sequence is highlighted in white bounding boxes along with corresponding predicted sentence in its bottom. The ground-truth sentences are presented on the right.
    }
	\label{demo} 
	\vspace{-0.2in}
\end{figure}

\begin{figure*}[]
	\centering
	\includegraphics[width=0.75\textwidth]{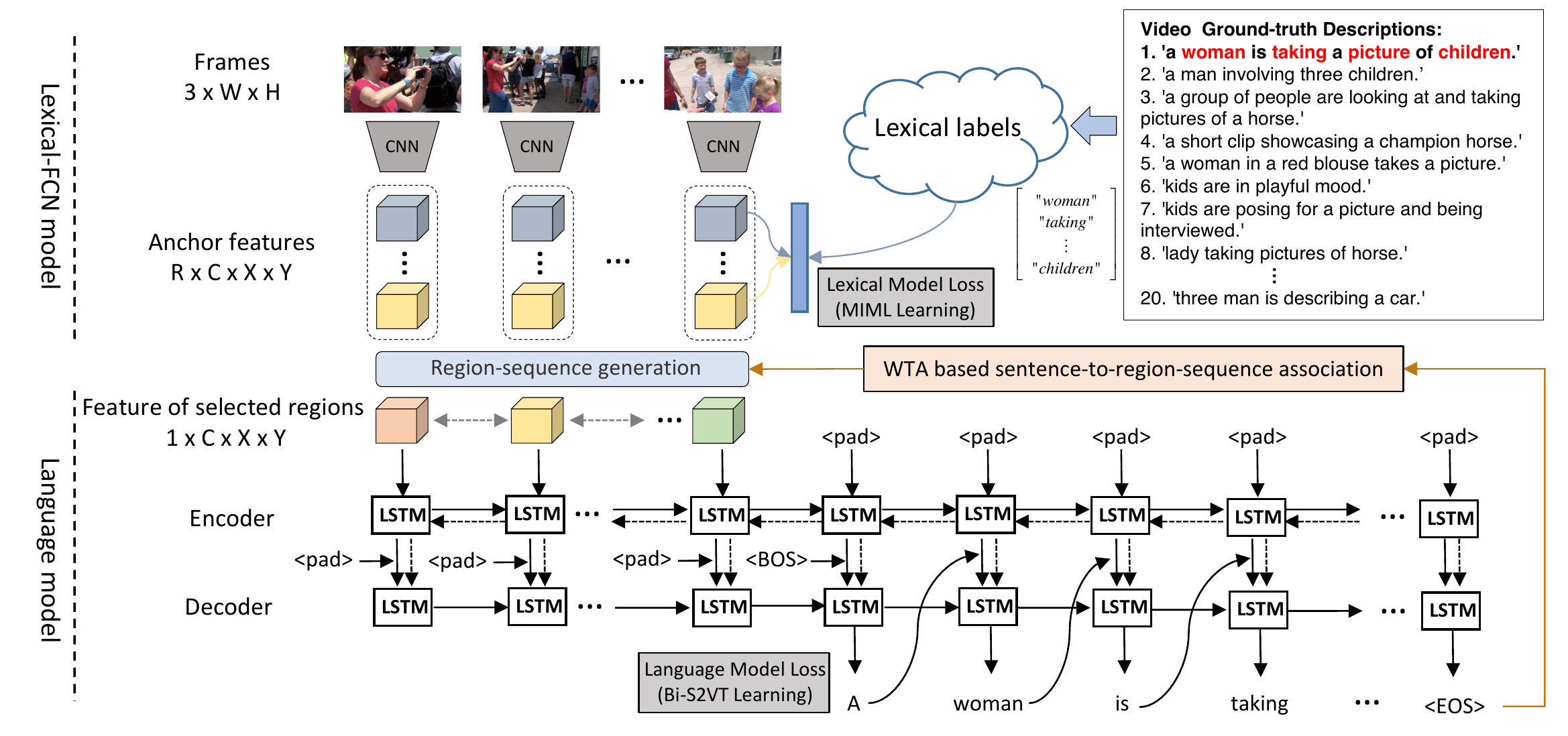}
	\vspace{-2 ex}
	\caption{Overview of our \textit{Dense Video Captioning} framework. In the language model, $<$BOS$>$ denotes the begin-of-sentence tag and $<$EOS$>$ denotes the end-of-sentence tag. We use zeros as $<$pad$>$ when there is no input at the time step. Best viewed in color.}
	\label{overview}
	\vspace{-0.2in}
\end{figure*}

However, using a single sentence cannot well describe the rich contents within images/videos.
The task of dense image captioning is therefore proposed, which aims to generate multiple sentences for different detected object locations in images \cite{johnson2015densecap,karpathy2015deep,kulkarni2013babytalk}. However, this setting requires region-level caption annotations for supervised training purpose.
As is well-known, videos are much more complex than images since the additional temporal dimension could provide informative contents such as different viewpoints of objects, object motions, procedural events, etc.
It is fairly expensive to provide region-sequence level sentence annotations for dense video captioning. The lack of such annotations has largely limited the much-needed progress of dense video captioning.
Our work in this paper is motivated by the following two questions.
First, most existing datasets have multiple video-level sentence annotations, which usually describe very diverse aspects (regions/segments) of the video clip.
However, existing video captioning methods simply represented all sentence descriptions with one global visual representation. This one-to-many mapping is far from accurate.
It is thus very interesting to investigate if there is an automatic way to (even weakly) associate sentence to region-sequence.
Second, is it possible to perform dense video captioning with those weakly associations (without
strong 1-to-1 mapping between sentences and region-sequence) in a weakly supervised fashion?

In this paper, we propose an approach to generate multiple diverse and informative captions
by weakly supervised learning from only the video-level sentence annotations.
Figure~\ref{overview} illustrates the architecture of the proposed approach, which consists of three major components: visual sub-model, region-sequence sub-model and language sub-model.
The visual sub-model is a lexical-FCN trained with weakly supervised multi-instance multi-label learning,
which builds the weak mapping between sentence lexical words and grid regions.
The second component solves the region-sequence generation problem.
We propose submodular maximization scheme to automatically generate informative and diverse region-sequences based on Lexical-FCN outputs.
A winner-takes-all scheme is proposed to weakly associate sentences to region-sequences in the training phase.
The third component generates sentence output for each region-sequence with a sequence-to-sequence learning based language model~\cite{venugopalan2015sequence}.
The main contributions are summarized as follows:
\begin{itemize}
\addtolength{\itemsep}{-0.05in}
\item[(1)] To the best of our knowledge, this is the first work for dense video captioning with only video-level sentence annotations.
\item[(2)] We propose a novel dense video captioning approach,
which models visual cues with Lexical-FCN, discovers region-sequence with submodular maximization, and decodes language outputs with sequence-to-sequence learning. Although the approach is trained with weakly supervised signal,
we show that \textit{informative} and \textit{diverse} captions can be produced.
\item[(3)] We evaluate dense captioning results by measuring the performance gap to oracle results,
 and diversity of the dense captions. The results clearly verify the advantages of the proposed approach.
   Especially, the best single caption by the proposed approach outperforms the state-of-the-art results on the MSR-VTT challenge by a large margin.
\end{itemize}

\section{Related Work}
\textbf{Multi-sentence description for videos} has been explored in various works recently~\cite{rohrbach2014coherent,shin2016beyond,yu2015video,das2013thousand,khan2011human}. Most of these works~\cite{yu2015video,shin2016beyond,rohrbach2014coherent} focused on generating a long caption (story-like), which first temporally segmented the video with action localization~\cite{shin2016beyond} or different levels of details~\cite{rohrbach2014coherent}, and then generated multiple captions for those segments and connected them with natural language processing techniques.
However, these methods simply considered the temporally segmentation, and ignored the frame-level region attention and the motion-sequence of region-level objects.
%% since most annotated natural sentences only describe the object region intra-frames.
Yu {\em et al.}~\cite{yu2015video} considered both the temporal and spatial attention, but still ignored the association or alignment of the sentences and visual locations.
In contrast, this paper tries to exploit both the temporal and spatial region information and further explores the correspondence between sentences and region-sequences for more accurate modeling.

\textbf{Lexical based CNN model} is of great advantages over the ImageNet based CNN model~\cite{russakovsky2015imagenet} in image/video captioning,
since the ImageNet based CNN model only captures a limited number of object concepts,
while the lexical based CNN model is able to capture all kinds of semantic concepts (nouns for objects and scenes, adjective for shape and attributes, verb for actions, etc).
It is non-trivial to  adopt/fine-tune the existing ImageNet CNN models with lexical output.
Previous works~\cite{fang2015captions,anne2016deep,venugopalan2016captioning,rohrbach2015long,kulkarni2013babytalk} have proposed several ways for this purpose.
For instance,~\cite{fang2015captions} adopted a weakly supervised multiple instance learning (MIL) approach~\cite{maron1998framework,zhang2005multiple} to train a CNN based word detector without the annotations of image-region to words correspondence;
and \cite{anne2016deep} applied a multiple label learning (MLL) method to learn the CNN based mapping between visual inputs and multiple concept tags.

\textbf{Sequence to sequence learning} with long short-term memory (LSTM)~\cite{hochreiter1997long} was initially proposed in the field of machine translation~\cite{sutskever2014sequence}.
Venugopalan {\em et al.} (S2VT)~\cite{venugopalan2015sequence} generalized it to video captioning.
Compared with contemporaneous works~\cite{yao2015describing,xu2016msr,pan2015jointly} which require additional temporal features from 3D ConvNets~\cite{tran2015learning}, S2VT can directly encode the temporal information by using LSTM on the frame sequence, and no longer needs the frame-level soft-attention mechanism~\cite{yao2015describing}.
This paper adopts the S2VT model~\cite{venugopalan2015sequence} with a bi-directional formulation to improve the encoder quality,
which shows better performance than the vanilla S2VT model in our experiments.
\begin{figure}[]
	\centering
	\includegraphics[width=0.72\linewidth]{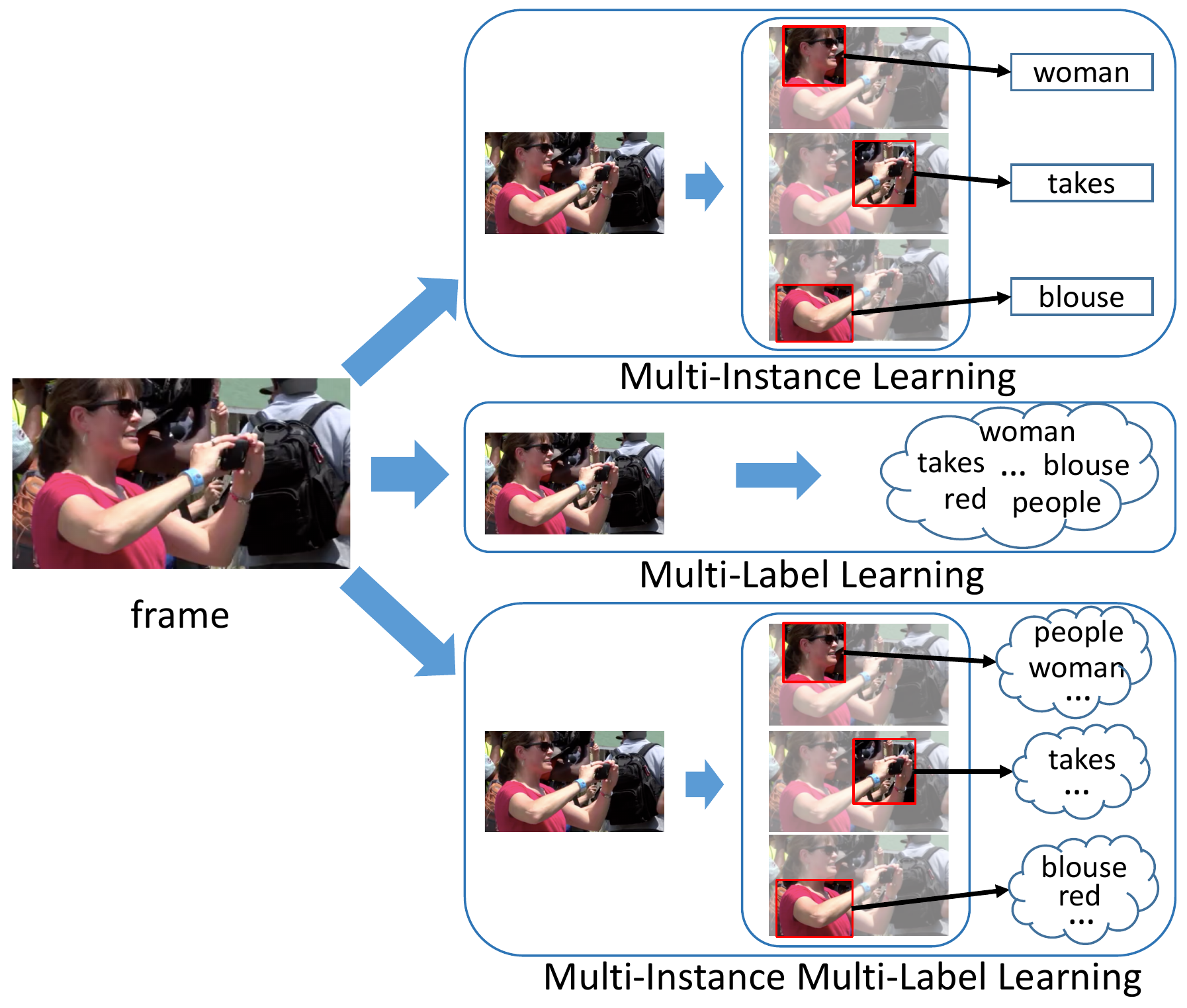}
    \vspace{-1.5 ex}
	\caption{Three paradigms of learning a lexical model. }	
	\label{MIMLL} 
	\vspace{-0.15in}
\end{figure}

\section{Approach}
Our ultimate goal is to build a system that describes input videos with dense caption sentences.
The challenges are two folds.
First, we do not have fine-grained training-data annotations which link sentence captions to region-sequences.
Second, we must ensure the generated sentences being informative and diverse. As discussed earlier, the proposed approach consists of three components (see Figure~\ref{overview}): lexical-FCN based visual model, region-sequence generation and language model. We elaborate each of them in the following.

\subsection{Lexical FCN Model}\label{MIMLL-FCN}
\subsubsection{Multi-instance Multi-label Lexical Model}\label{preli}
We adopt multi-instance multi-label learning (MIMLL) to train our lexical model, which could be viewed as a combination of word detection~\cite{fang2015captions} (MIL) and deep lexical classification~\cite{anne2016deep} (MLL). Figure~\ref{MIMLL} illustrates the comparison of the three methods.

\textbf{Multi-instance learning}~\cite{maron1998framework,zhang2005multiple,fang2015captions} assumes that the word label $y_i^w$ is assigned to a bag of instances $\mathbf{X}_i=\{\mathbf{x}_{i1},\dots,\mathbf{x}_{ij}\}$ where $i$ is the bag index,
${\mathbf{x}_{ij}} \in {\mathbb{R}^d}$ is a $d$-dimensional feature vector for the $j$-th instance.
The word detection method~\cite{fang2015captions} used fc7 features of VGG-16 as the instance representations. The bag is positive with a word label $y_i^w=1$ if at least one of the instances in $\mathbf{X}_i$ contains the word $w$, although it is not exactly known which one contains the word. The bag is negative with label $y_i^w=0$ if no instance contains the word $w$.

\textbf{Multi-label learning} assumes that each instance $\mathbf{x}_i$ has multiple word labels: $\mathbf{y}_i=\{y_i^1,\dots,y_i^{\mathbf{k}}\}$ where $\mathbf{k}$ is the number of labels. For this purpose, we usually train a deep neural network with a sigmoid cross-entropy loss \cite{anne2016deep}.

\textbf{Multi-instance multi-label learning}~\cite{zhou2006multi} is a natural generalization of MIL. It takes as input pairs $\{\mathbf{X}_i,\mathbf{y}_i\}$, where each $\mathbf{X}_i$ is a bag of instances labeled with a set of words $\mathbf{y}_i=\{y_i^1,\dots,y_i^{\mathbf{k}}\}$. In MIMLL, each instance usually has one or multiple word labels. For instance, we can use ``woman'', ``people'', ``human'' or other synonyms in the lexicon to describe a female (see Figure~\ref{MIMLL} for one example).
Now we define the loss function for a bag of instances. As each bag has multiple word labels, we adopt the cross-entropy loss to measure the multi-label errors:
\begin{equation}
\footnotesize
    L(\mathbf{X},\mathbf{y};\theta ) =  - \frac{1}{N}\sum\limits_{i = 1}^N {\left[ {\mathbf{y}{_i} \cdot \log {\mathbf{\hat p}_i}
    + (1 - \mathbf {y}{_i}) \cdot \log (1 - {\mathbf{\hat p}_i})} \right]},
\end{equation}
where $\theta$ is the model parameters, $N$ is the number of bags, $\mathbf{y}_i$ is the label vector for bag $\mathbf{X}_i$, and $\mathbf{\hat p}_i$ is the corresponding probability vector. We weakly label the bag as negative when all instances in the bag are negative, and thus use a noisy-OR formulation~\cite{heckerman2013tractable,maron1998framework} to combine the probabilities that the individual instances in the bag are negative:
\begin{equation}
\footnotesize
{\hat p_i}^w = P(y_i^w = 1|{\mathbf{X}_i};\theta ) = 1 - \prod\limits_{\mathbf{x}_{ij} \in {\mathbf{X}_i}} {(1 - P(y_i^w = 1|{\mathbf{x}_{ij}};\theta ))},
\end{equation}
where ${\hat p_i}^w$ is the probability when word $w$ in the $i$-th bag is positive.
We define a sigmoid function to model the individual word probability:
\begin{equation}
\footnotesize
 P(y_{i}^w = 1|{\mathbf{x}_{ij}};\theta) = \sigma ({\mathbf{w}_w}{\mathbf{x}_{ij}} + {\mathbf{b}_w}),
 \label{probability}
\end{equation}
where $\mathbf{w}_w$ is the weight matrices, $\mathbf{b}_w$ is the bias vector,
and $\sigma(x) = 1/( 1 + \exp(-x))$ is the logistic function.
In our Lexical-FCN model, we use the last pooling layer (pool5 for ResNet-50) as the representation of instance $\mathbf{x}_{ij}$,
which will be elaborated in the following sections.

\subsubsection{Details of Lexical-FCN}\label{FCN}
Lexical-FCN model builds the mapping between frame regions and lexical labels.
The first step of Lexical-FCN is to build a lexical vocabulary from the video caption training set.
We extract the part-of-speech~\cite{toutanova2003feature} of each word in the entire training dataset.
These words may belong to any part of sentences, including nouns, verbs, adjectives and pronouns.
We treat some of the most frequent functional words\footnote{Functional words are `is', `are', `at', `on', `in', `with', `and' and `to'.} as stop words, and remove them from the lexical vocabulary.
We keep those remaining words appearing at least five times in the MSR-VTT training set, and finally obtain a vocabulary $\mathcal{V}$ with 6,690 words.

The second step of Lexical-FCN is to train the CNN models with MIMLL loss described above.
Instead of training from scratch, we start from some state-of-the-art ImageNet models like VGG-16~\cite{simonyan2014very} or ResNet-50~\cite{he2015deep}, and fine-tune them with the MIMLL loss on the MS-VTT training set.
For VGG-16, we re-cast the fully connected layers to convolutions layers to obtain a FCN.
For ResNet-50, we remove final softmax layer and keep the last mean pooling layer to obtain a FCN.
%Given an input frame of size $3 \times W \times H$, the Lexical-FCN will output a $C \times \left( {\left\lfloor {\frac{W}{{32}}} \right\rfloor  - 6} \right) \times \left( {\left\lfloor {\frac{H}{{32}}} \right\rfloor  - 6} \right)$ response maps ($C = 4096$ for VGG-16 and $2048$ for ResNet-50).

\subsubsection{Regions from Convolutional Anchors}\label{region}
In order to obtain the dense captions, we need grounding the sentences to sequences of ROI (regions of interest).
Early solutions in object detection adopt region proposal algorithms to generate region candidates,
and train a CNN model with an additional ROI pooling layer~\cite{he2014spatial,girshick2015fast,ren2015faster}.
This cannot be adopted in our case, since we do not have the bounding box ground-truth for any words or concepts required in the training procedure.
Instead, we borrow the idea from YOLO~\cite{yolo2016}, and generate coarse region candidates from anchor points of the last FCN layer \cite{long2015fully,fang2015captions}.
In both training and inference phases, we sample the video frames and resize both dimensions to 320 pixels.
After feeding forward through the FCN, we get a 4$\times$4 response map (4096 channels for VGG-16 and 2048 channels for ResNet-50).
Each anchor point in the response map represents a region in the original frame. Unlike object detection approaches, the bounding-box regression process is not performed here since we do not have the ground-truth bounding boxes.
We consider the informative region-sequence generation problem directly starting with these 16 very-coarse grid regions.

\subsection{Region-Sequence Generation}\label{feat}
Regions between different frames are matched and connected sequentially to produce \textit{region-sequences}.
As each frame has 16 coarse regions, even if each video clip is downsampled to 30 frames, we have to face a search space of size $16^{30}$ for region-sequence generation.
This is intractable for common methods even for the training case that has video-level sentence annotations.
However, our Lexical-FCN model provides the lexical descriptions for each region at every frame, so that we can consider the problem from a different perspective.

\subsubsection{Problem Formulation}
We formulate the region-sequence generation task as a sub-set selection problem~\cite{leskovec2007cost,gygli2015video},
in which we start from an empty set, and sequentially add one most informative and coherent region at each frame into the subset,
and in the meantime ensure the diversity among different region-sequences.
Let $\mathcal{S}_{\mathbf{v}}$ denote the set of all possible region sequences of video $\mathbf{v}$,
$\mathcal{A}$ is a region-sequence sub-set, i.e., $\mathcal{A}  \subseteq \mathcal{S}_{\mathbf{v}} $.
Our goal is to select a region-sequence $\mathcal{A}^*$, which optimizes an objective ${R}$:
\begin{equation}
\footnotesize
{\mathcal{A}^*} = \arg \max\limits_{\mathcal{A} \subseteq {\mathcal{S}_\mathbf{v}}} \; {R}({\mathbf{x_v}},\mathcal{A}),
\label{eq:obj}
\end{equation}
where $\mathbf{x_v}$ are all region feature representations of video $\mathbf{v}$.
We define ${R}({\mathbf{x_v}},\mathcal{A})$ as linear combination objectives
\begin{equation}
\footnotesize
{R}({\mathbf{x_v}},\mathcal{A}) = {\mathbf{w_v}^T} \mathbf{f}({\mathbf{x_v}},\mathcal{A}),
\end{equation}
where $\mathbf{f} = [f_{inf}, f_{div}, f_{coh}]^T$, which describe three aspects of the region-sequence, i.e., {\em informative}, {\em diverse} and {\em coherent}.
The optimization problem of Eq-\ref{eq:obj} quickly becomes intractable when $\mathcal{S}_{\mathbf{v}}$ grows exponentially with the video length.
We restrict the objectives $\mathbf{f}$ to be monotone submodular function and $\mathbf{w_v}$ to be non-negative.
This allows us to find a near optimal solution in an efficient way.

\subsubsection{Submodular Maximization}
We briefly introduce submodular maximization and show how to learn the weights $\mathbf{w_v}$.
A set function is called submodular if it fulfills the \textit{diminishing returns} property.
That means, given a function $f$ and arbitrary sets $\mathcal{A} \subseteq \mathcal{B} \subseteq \mathcal{S}_\mathbf{v}\setminus{r}$,
$f$ is submodular if it satisfies:
\begin{equation}
\footnotesize
f(\mathcal{A} \cup \{r\})- f(\mathcal{A}) \geq f(\mathcal{B}\cup \{r\}) - f(\mathcal{B}).
\end{equation}
Linear combination of submodular functions is still submodular for non-negative weights.
For more details, please refer to \cite{nemhauser1978analysis,leskovec2007cost}.

Submodular functions have many properties that are similar to convex or concave functions, which are desirable for optimization.
Previous works~\cite{nemhauser1978analysis,leskovec2007cost,gygli2015video} have shown that maximizing a submodular function with a greedy algorithm yields a good approximation to the optimal solution.
In this paper, we apply a commonly used cost-effective lazy forward (CELF) method~\cite{leskovec2007cost} for our purpose.
We defined a marginal gain function as
\begin{equation}
\footnotesize
\begin{split}
\mathcal{L}(\mathbf{w_v}; r) & = R({\mathcal{A}_{\mathbf{t} - 1}} \cup \{ r \})  - R({\mathcal{A}_{\mathbf{t} - 1}}) \\
 & = {\mathbf{w_v}^T}\mathbf{f}({\mathbf{x_v}},{\mathcal{A}_{\mathbf{t} - 1}} \cup \{ r \})
    - {\mathbf{w_v}^T}\mathbf{f}({\mathbf{x_v}},{\mathcal{A}_{\mathbf{t} - 1}}).
\end{split}
\end{equation}
The CELF algorithm starts with an empty sequence $\mathcal{A}_0=\emptyset$, and adds the region $r_t$ at step $t$ into region-sequence
which can maximize the marginal gain:
\begin{equation}
\footnotesize
\mathcal{A}_{\mathbf{t}} = \mathcal{A}_{\mathbf{t} -1 } \cup \{r_t\}; ~~
 {r_t} = \arg \max\limits_{r \in {\mathcal{S}_t}} \mathcal{L}(\mathbf{w_v}; r),
\end{equation}
where ${\mathcal{S}_t}$ means region sets in frame-$t$.

Given $N$ pairs of known correspondences $\{(\mathbf{r}, \mathbf{s})\}$, we optimize $\mathbf{w_v}$ with the following objective:
\begin{equation}
\footnotesize
    \min\limits_{\mathbf{w_v} \geq 0} \frac{1}{N} \sum\limits_{i=1}^N { \max\limits_{r \in \mathbf{r}_i}  \mathcal{L}_i(\mathbf{w_v}; r) + \frac{\lambda}{2} \|\mathbf{w_v}\|^2},
\end{equation}
where the max-term is a generalized hinge loss, which means ground-truth or oracle selected region $r$ should have a higher score than any other regions by some margin.

Our training data do not have $(\mathbf{r}, \mathbf{s})$ pairs, i.e., the sentence to region-sequence correspondence.
We solve this problem in a way that is similar to the alternative directional optimization:
(1) we initialize $\mathbf{w_v}$ = $\mathbf{1}$ (all elements equals to 1);
(2) we obtain a region-sequence with submodular maximization with that $\mathbf{w_v}$;
(3) we weakly associate sentence to region-sequence with a winner-takes-all (WTA) scheme (described later);
(4) we refine $\mathbf{w_v}$ with the obtained sentence to region-sequence correspondence;
(5) we repeat step-2$\sim$4 until $\mathbf{w_v}$ is converged.

The WTA scheme works in four steps when giving a ground-truth sentence $\mathbf{s}$.
\textit{First}, we extract the lexical labels from $\mathbf{s}$ based on the vocabulary $\mathcal{V}$, and form a lexical subset $\mathcal{V}_\mathbf{s}$.
\textit{Second}, we obtain probability of word $w\in \mathcal{V}_\mathbf{s}$ for the $i$-th region-sequence by
$p_i^w = \max\nolimits_{j} p_{ij}^w$, where $p_{ij}^w$ is the probability of word $w$ in the $j$-th frame, which is in fact from the Lexical-FCN output for each region.
\textit{Third}, we threshold $p_i^w$ with a threshold $\theta$, i.e., redefining $p_i^w = 0$ if $p_i^w < \theta$ ($\theta$ = 0.1 in our studies).
\textit{Last}, we compute the matching score by
\begin{equation}
\footnotesize
    f_i = \sum\limits_{w \in \mathcal{V}_\mathbf{s};~p_i^w \geq \theta } p_i^w,
\label{eq:finfsup}
\end{equation}
and obtain the best region-sequence by $i^\ast = \arg \max\nolimits_{i} f_i$.
This objective suggests that we should generate region-sequences having high-scored words in the sentences.
\begin{figure}[]
	\centering
	\includegraphics[width=0.74\linewidth]{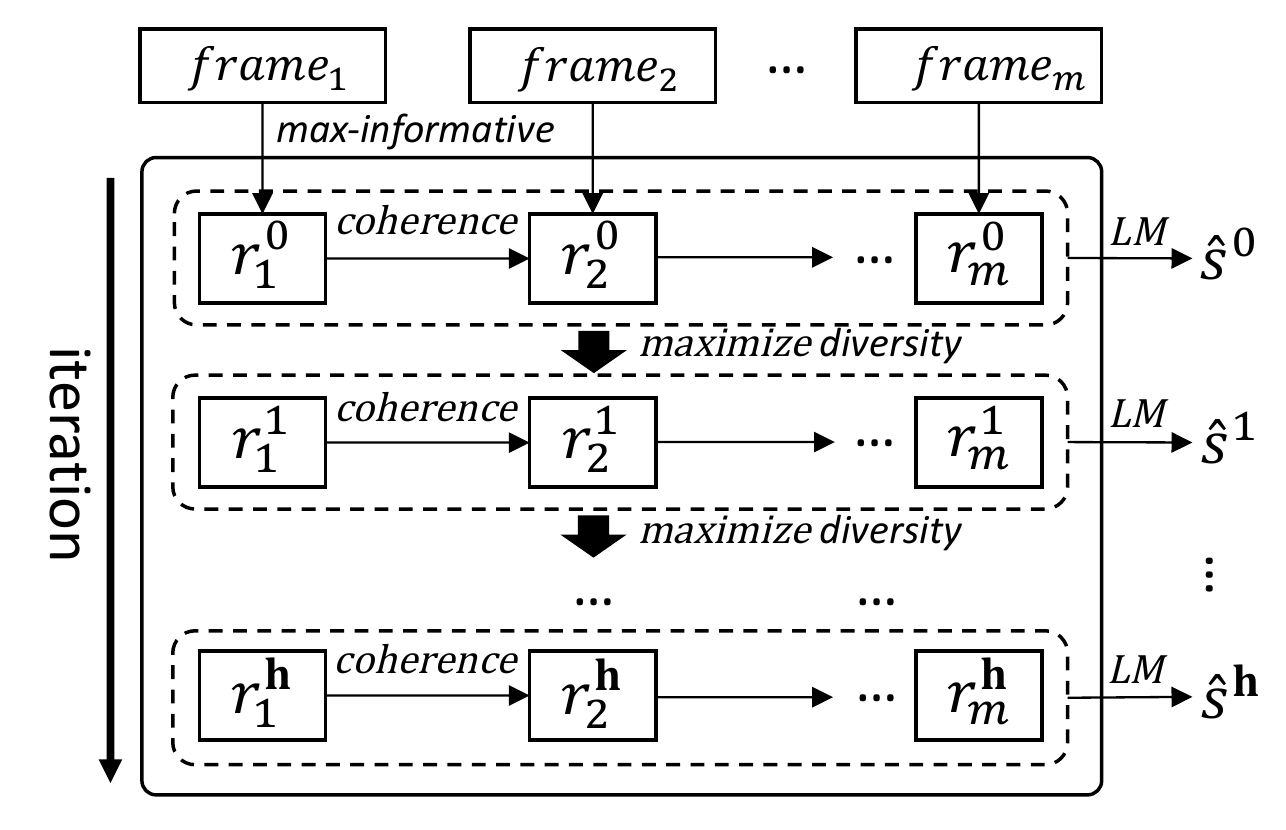}
    \vspace{-2 ex}
	\caption{Illustration of region-sequence generation. $r_i^j$ is the $j$-th region-sequence in $i$-th frame and `{\em {LM}}' denotes language model.}
	\label{inference}
	\vspace{-0.17in}
\end{figure}

\subsubsection{Submodular Functions}
Based on the properties of submodular function~\cite{lovasz1983submodular,nemhauser1978analysis}, we describe how to define the three components as follows.

\textbf{Informativeness} of a region-sequence is defined as the sum of each region's informativeness:
\begin{equation}
\footnotesize
{f_{\inf }}({\mathbf{x_v}},\mathcal{A}_t) = \sum\nolimits_{w} p^w; ~~~p^w = \max\limits_{i \in \mathcal{A}_t} p_{i}^w.
\label{eq:finf}
\end{equation}
If video-level sentence annotations are known either in the training case or by an oracle, we replace the definition with
Eq-\ref{eq:finfsup}, which limits words by the sentence vocabulary $\mathcal{V}_\mathbf{s}$.

\textbf{Coherence} aims to ensure the temporal coherence of the region-sequence, since significant changes of region contents may confuse the language model.
Similar to some works in visual tracking \cite{babenko2009visual,jepson2003robust},
we try to select regions with the smallest changes temporally, and define the coherence component as
\begin{equation}
\footnotesize
{f_{coh}} = \sum\nolimits_{r_s \in \mathcal{A}_{t-1}} \langle \mathbf{x}_{r_t} , \mathbf{x}_{r_s}\rangle,
\end{equation}
where $\mathbf{x}_{r_t}$ is the feature of region $r_t$ at $t$-th step, $\mathbf{x}_{r_s}$ is one of the region feature in the previous $(t-1)$ steps,
and $\langle, \rangle$ means dot-production operation between two normalized feature vectors.
In practice, we also limit the search space of region $r_t$ within the 9  neighborhood positions of the region from the previous step.

\textbf{Diversity} measures the degree of difference between a candidate region-sequence and all the existing region-sequences.
Suppose $\{{p}_i^w\}_{i=1}^N$ are the probability distribution of the existing $N$ region-sequences and ${q}^w$ is the probability distribution of a candidate region-sequence,
the diversity is defined with the Kullback-Leibler divergence as
\begin{equation}
\footnotesize
{f_{div}} = \sum_{i=1}^N \int\nolimits_{w} {p}_i^w \log \frac{{p}_i^w}{{q}^w} dw.
\end{equation}

We initially pick the most informative region-sequence, and feed it to a language model ({\em {LM}}) for sentence output.
Then we iteratively pick a region-sequence which maximizes diversity to generate multiple sentence outputs.
Figure \ref{inference} illustrates our region-sequence generation method.
The detailed algorithm is given in the supplementary file.

\subsection{Language Models}\label{language}
We model the weakly associated temporal structure between region-sequence and sentence with the sequence-to-sequence learning framework (S2VT) \cite{venugopalan2015sequence}, which is an encoder-decoder structure.
S2VT encodes visual feature of region-sequences $\vec{V}=(\mathbf{v_1}, \cdots , \mathbf{v_T})$ with LSTM,
and decodes the visual representation into a sequence of output words $\vec{u}=({u_1}, \cdots ,{u_S})$.
LSTM is used to model a sequence in both the encoder part and the decoder part. As a variant of RNN, LSTM is able to learn long-term temporal information and dependencies that traditional RNN is difficult to capture~\cite{hochreiter1997long}.
Our LSTM implementation is based on~\cite{zaremba2014recurrent} with dropout regularization on all LSTM units (dropout ratio 0.9).

We extend the original S2VT with bi-directional encoder, so that the S2VT learning in Figure~\ref{overview} stacks three LSTM models.
The first LSTM encodes forward visual feature sequence $\{\vec{V}\}$, and the second encodes the reverse visual feature sequence $\{\cev{V}\}$. These two LSTM networks form the encoder part.
We will show the benefit of bi-direction LSTM encoding later.
The third LSTM decodes visual codes from both the forward pass and backward pass into sequences of words (sentences).

To further improve accuracy, we propose a category-wise language model extension.
Videos may belong to different categories, such as news, sports, etc.
Different video category has very different visual patterns and sentence styles.
The category-wise language model is defined as
\begin{equation}
   \footnotesize
    \mathbf{s}^{\ast} = \arg\max\nolimits_{\mathbf{s}} P(s|c, \mathbf{v}) P(c|\mathbf{v}),
    \label{eq:cat}
\end{equation}
where $c$ is the category label, $\mathbf{v}$ is the video feature representation, and $\mathbf{s}$ is the predicted sentence.
$P(\mathbf{s}|c, \mathbf{v})$ is the probability conditional on category $c$ and video $\mathbf{v}$,
and $P(c|\mathbf{v})$ is prior confidence of video $\mathbf{v}$ belongs to a category $c$, which can be obtained from a general video categorization model.
The category-wise language model can be viewed as max-a-posterior estimation.
\begin{figure}[]
	\centering
	\includegraphics[width=0.75\linewidth]{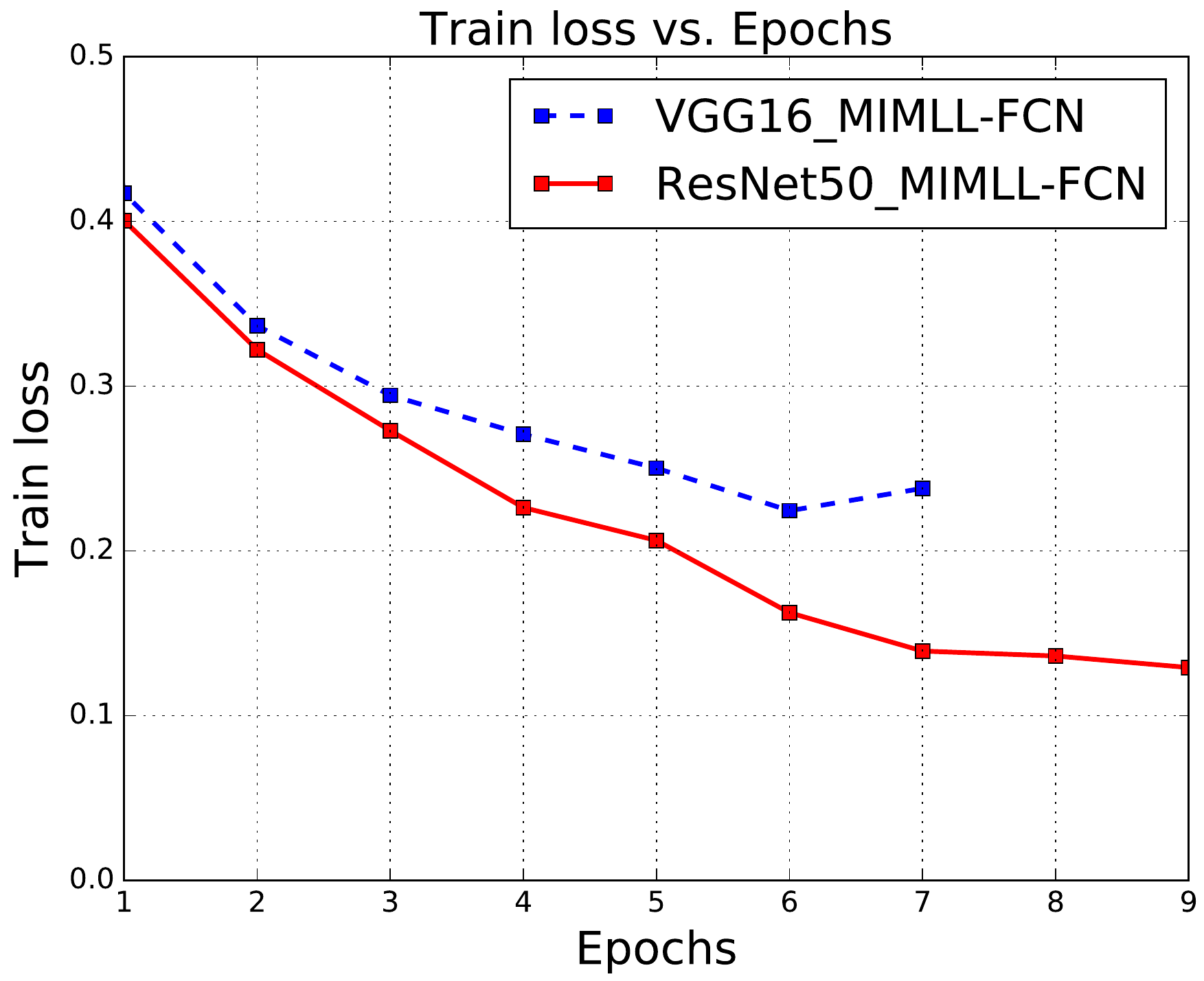}
    \vspace{-1.5 ex}
	\caption{The lexical training loss on the MSR-VTT dataset.}
	\label{curves} 
	\vspace{-0.15in}
\end{figure}

\section{Experiments}\label{exp}
\subsection{Dataset and Implementation Details}
We conduct experiments on the MSR-VTT dataset~\cite{xu2016msr}, which is a recently released large-scale video caption benchmark. This dataset contains 10,000 video clips (6,513 for training, 497 for validation and 2,990 for testing) from 20 categories, including news, sports, etc.
Each video clip is manually annotated with 20 natural sentences.
This is currently the largest video captioning dataset in terms of the amounts of sentences and the size of the vocabulary.
Although this dataset was mainly used for evaluating single sentence captioning results,
we assume that the 20 sentences for each clip contain very diversified annotations and can be used in the task of dense captioning (with some redundancy as will be discussed later).

For the evaluation of single captioning, the authors of this benchmark proposed machine translation based metrics like METEOR~\cite{lavie2014meteor}, BLEU@1-4~\cite{papineni2002bleu}, ROUGE-L~\cite{lin2004rouge} and CIDEr~\cite{vedantam2015cider}.
For dense video captioning results, we propose our own evaluation protocol to justify the results.

All the training and testing are done on an Nvidia TitanX GPU with 12GB memory.
Our model is efficient during the testing stage.
It can process a 30-frame video clip in about 840ms on the TitanX GPU, including 570ms for CNN feature extraction, 90ms for region-sequence generation, and 180ms for language model.

\subsection{Ablation Studies on Single Sentence Captioning}
We first evaluate the effect of several design components through single sentence captioning experiments, which
produce a caption with the maximal informative score defined by Eq-\ref{eq:finf} (i.e., $\hat{\mathbf{s}}^0$ in Figure \ref{inference}).

\textbf{Effectiveness of Network Structure}.
We compare VGG-16 and ResNet-50 for the Lexical-FCN model. Due to the GPU memory limitation, we do not try a deeper network like ResNet-152.
Figure~\ref{curves} shows that ResNet-50 achieves better training loss than VGG-16, which is consistent with their results on ImageNet.
Table~\ref{compare} summarizes the single sentence captioning results on the MSR-VTT validation set by two networks.
As can be seen, in all the cases, ResNet-50 performs better than VGG-16.
Based on these results, we choose ResNet-50 as our network structure in the following studies when there is no explicit statement.

\textbf{Effectiveness of Bi-directional Encoder}.
Next we compare the performances of \textit{bi-directional} and \textit{unidirectional} S2VT models for language modeling.
Results are also shown in Table~\ref{compare}.
It is obvious that \textit{bi-directional} model outperforms \textit{unidirectional} model on all the evaluated metrics.
The benefit of \textit{bi-directional} model is not that significant.
We conjecture that this is due to the fact that the region-sequences already include enough temporal and local information.
Nevertheless, for better accuracy, all the following studies adopt the \textit{bi-directional} model.
\begin{table}[]
	\centering
    \small
	\resizebox{0.95\linewidth}{!}{
		\begin{tabular}{l|c|c|c|c}
			\hline
			\textbf{Model}  & \multicolumn{1}{l|}{\textbf{METEOR}} & \multicolumn{1}{l|}{\textbf{BLEU@4}} & \multicolumn{1}{l|}{\textbf{ROUGE-L}} & \multicolumn{1}{l}{\textbf{CIDEr}} \\ \hline			
			\textbf{Unidirectional (VGG-16)}     &   25.2   &    32.7    &  56.0   &   31.1    \\
			\textbf{Bi-directional (VGG-16)}     &    25.4  &  32.8      &   56.5  &   32.9    \\
			\textbf{Unidirectional (ResNet-50)}  &    25.7  &   32.1     & 56.4    &  32.5     \\
			\textbf{Bi-directional (ResNet-50)}  &   25.9   &   33.7     &  56.9   &   32.6    \\ \hline
		\end{tabular}
	}
    \vspace{-1.5 ex}
	\caption{Single sentence captioning accuracy by bi-/uni-directional encoder on the {\textit{validation set}} of MSR-VTT.
	}
	\label{compare}
	\vspace{-0.1in}
\end{table}
\begin{table}[]
	\centering
    \small
	\resizebox{0.95\linewidth}{!}{
		\begin{tabular}{l|c|c|c|c}
			\hline
			\textbf{Model}    & \multicolumn{1}{l|}{\textbf{METEOR}} & \multicolumn{1}{l|}{\textbf{BLEU@4}} & \multicolumn{1}{l|}{\textbf{ROUGE-L}} & \multicolumn{1}{l}{\textbf{CIDEr}} \\ \hline
			\textbf{MIL (bi-directional)}   &  23.3 & 28.7  & 53.1  & 24.4  \\
			\textbf{MIMLL (bi-directional)} & 25.9 & 33.7 & 56.9 & 32.6 \\ \hline
		\end{tabular}
	}
    \vspace{-1.5 ex}
	\caption{Single sentence captioning accuracy by MIL and MIMLL on the {\textit{validation set}} of MSR-VTT.}
	\label{compare_MIL_MIML}
	\vspace{-0.15in}
\end{table}

\textbf{Effectiveness of MIMLL}.
Our Lexical-FCN model is trained on video frames.
Compared with image-level lexical learning~\cite{fang2015captions,anne2016deep},
our setting is much more challenging since the sentences are on the video-level, and it is hard to determine which words correspond to which frames.
Here we show the effectiveness of the MIMLL in two aspects.
First, we compare the single captioning results by MIMLL and MIL in Table~\ref{compare_MIL_MIML}.
We can see that MIMLL achieves better accuracy than MIL on all the four metrics.
Second, we compare the word detection accuracy of MIMLL and MIL.
We first compute the max-probability of each word within the region-sequence.
If the max-probability of a word is greater than a threshold (0.5), we claim that the word is detected.
We observe that MIMLL is better in detecting accuracy than MIL in this study (43.1\% vs 41.3\%).
Both results demonstrate the effectiveness of the proposed MIMLL for the Lexical-FCN model.

\textbf{Effectiveness of Category-wise Language Model}.
All the previous studies are based on language model without using video category information.
Here, we study the benefit of the category-wise language model, as defined in Eq-\ref{eq:cat}.
Results are shown in the 2nd last and the 3rd last rows in Table~\ref{val_res}.
We observe that the category-wise language model achieves much better accuracy than that without category-wise modeling.
The benefit is due to that category information provides a strong prior about video content.

\textbf{Comparison with State-of-the-arts}.
We also compare our single sentence captioning results with the state-of-the-art methods in MSR-VTT benchmark.
For better accuracy, this experiment adopts data augmentation during the training procedure, similar to the compared methods.
We preprocess each video clip to 30-frames with different sampling strategies (random, uniform, etc), and obtain multiple instances for each video clip.

We first compare our method with mean-pooling~\cite{venugopalan2014translating}, soft-attention~\cite{yao2015describing} and S2VT~\cite{venugopalan2015sequence} on the validation set of MSR-VTT.
All these alternative methods have source codes available for easy evaluation. Results are summarized in Table~\ref{val_res}.
Our baseline approach (the 2nd last row) is significantly better than these 3 methods.
We also compare with the top-4 results from the MSR-VTT challenge in the table, including v2t\_navigator~\cite{v2t_navigator}, Aalto~\cite{shetty2016frame}, VideoLAB~\cite{VideoLAB} and ruc\_uva~\cite{dong2016early}\footnote{\url{http://ms-multimedia-challenge.com/}.},
which are all based on features from multiple cues such as action features like C3D and audio features like Bag-of-Audio-Words (BoAW)~\cite{pancoast2014softening}.
Our baseline has on-par accuracy to the state-of-the-art methods.
For fair comparison, we integrate C3D action features and audio features together with our lexical features and feed them into the language model. Clearly better results are observed.

In Table~\ref{test_res}, we compare our results on the test set of MSR-VTT with the top-4 submissions in the challenge leaderboard, where we can see that similar or better results are obtained in all the four evaluated metrics.
\begin{table}[]
	\centering
    \small
	\resizebox{0.95\linewidth}{!}{
		\begin{tabular}{l|c|c|c|c}
			\hline
			\textbf{Model}  & \multicolumn{1}{l|}{\textbf{METEOR}} & \multicolumn{1}{l|}{\textbf{BLEU@4}} & \multicolumn{1}{l|}{\textbf{ROUGE-L}} & \multicolumn{1}{l}{\textbf{CIDEr}} \\ \hline
			\textbf{Mean-Pooling~\cite{venugopalan2014translating}} &     23.7 &     30.4 &      52.0  &    35.0    \\
			\textbf{Soft-Attention~\cite{yao2015describing}}    &     25.0 &     28.5 &      53.3  &    37.1    \\
			\textbf{S2VT~\cite{venugopalan2015sequence}}    &     25.7 &     31.4 &      55.9  &    35.2    \\ \hline
			\textbf{{ruc-uva~\cite{dong2016early}}}        &     27.5 &     39.4 &      60.0  &    48.0    \\
			\textbf{{VideoLAB~\cite{VideoLAB}}}             &     27.7 &     39.5 &      61.0  &    44.2    \\
			\textbf{{Aalto~\cite{shetty2016frame}}}         &     27.7 &     41.1 &      59.6  &    46.4    \\
			\textbf{{v2t\_navigator~\cite{v2t_navigator}}}  &     29.0 &     43.7 &      61.4  &    45.7    \\  \hline
			%\textbf{{VideoLAB (committee)~\cite{VideoLAB}}} &     28.6 &     40.7 &      61.0  &    46.5    \\
            \textbf{Ours w/o category}   & 27.7 &     39.0 &      60.1  &    44.0    \\
			\textbf{Ours category-wise}           &     28.2 &     40.9 &      61.8  &    44.7    \\
			\textbf{Ours + C3D + Audio}                    &  \textbf{29.4} &\textbf{44.2} &      \textbf{62.6}  &     \textbf{50.5} \\ \hline
		\end{tabular}
	}
    \vspace{-1 ex}
	\caption{Comparison with state of the arts on the {\textit{validation set}} of MSR-VTT dataset. See texts for more explanations.}
	\label{val_res}
\vspace{-0.1in}
\end{table}

\begin{table}[]
	\centering
     \small
	\resizebox{0.95\linewidth}{!}{
		\begin{tabular}{l|c|c|c|c}
			\hline
			\textbf{Model}   & \multicolumn{1}{l|}{\textbf{METEOR}} & \multicolumn{1}{l|}{\textbf{BLEU@4}} & \multicolumn{1}{l|}{\textbf{ROUGE-L}} & \multicolumn{1}{l}{\textbf{CIDEr}} \\ \hline
			\textbf{{ruc-uva~\cite{dong2016early}}}        &     26.9 &     38.7 &      58.7  &    45.9    \\
			\textbf{{VideoLAB~\cite{VideoLAB}}}            &     27.7 &     39.1 &      60.6  &    44.1    \\
			\textbf{{Aalto~\cite{shetty2016frame}}}        &     26.9 &     39.8 &      59.8  &    45.7    \\
			\textbf{{v2t\_navigator~\cite{v2t_navigator}}} &     28.2 &     40.8 &      60.9  &    44.8    \\ \hline
			% \textbf{{Fudan-ILC}}                           &     26.8 &     38.7 &      59.5  &    41.9    \\
			\textbf{{ Ours}}                  &  \textbf{28.3}   & \textbf{41.4}   &    \textbf{61.1}   &   \textbf{48.9}     \\ \hline
		\end{tabular}
	}
    \vspace{-1 ex}
	\caption{Comparison with state of the arts on the \textit{test set} of MSR-VTT dataset. See texts for more explanations.}
	\label{test_res}
   \vspace{-0.15in}
\end{table}

\subsection{Evaluation of Dense Captioning Results} \label{DenseDescription}
The proposed approach can produce a set of region-sequences with corresponding multiple captions for an input video clip.
Besides qualitative results in Figure~\ref{demo} and the supplementary file, we evaluate the results quantitatively in two aspects:
1) performance gap between automatic results and oracle results, and
2) diversity of the dense captions.

\subsubsection{Performance Gap with Oracle Results}
We measure the quality of dense captioning results by the performance gap between our automatic results and oracle results.
Oracle leverages information from ground-truth sentences to produce the caption results.
Oracle information could be incorporated in two settings.
\textit{First}, similar to the training phase, during inference oracle uses the ground-truth information to guide sentence to region-sequence association.
\textit{Second}, oracle uses the ground-truth sentences to measure the goodness of each caption sentence using metrics like METEOR and CIDEr, and re-ranks the sentences according to their evaluation scores.
It is obvious that the oracle results are the upper bound of the automatic method.

Inspired by the evaluation of dense image captioning \cite{johnson2015densecap},
we use averaged precision (AP) to measure the accuracy of dense video captioning.
We compute the precision in terms of all the four metrics (METEOR, BLEU@4, ROUGE-L and CIDEr) for every predicted sentence,
and obtain average values of the top-5 and top-10 predicted sentences.
The gap of AP values between oracle results and our automatic results will directly reflect the effectiveness of the automatic method.

For our automatic method, the output sentences need to be ranked to obtain the top-5 or top-10 sentences.
Similar to \cite{shetty2016frame},
we train an evaluator network in a supervised way for this purpose, since
submodular maximization does not ensure that sentences are generated in quality decreasing order.
Table~\ref{top_res} lists the comparative results on the validation set of MSR-VTT using three strategies:
(1) oracle for both sentence to region-sequence association and sentence re-ranking (OSR + ORE in short);
(2) dense video captioning + oracle re-ranking (Dense + ORE in short);
(3) fully automatic dense video captioning method (DenseVidCap).

Results indicate that the dense video captioning + oracle re-ranking could
reach $\geq$95\% relative accuracy of the ``fully" oracle results (OSR+ORE) on all the metrics for the top-5 sentences, and
 $\geq$93\% relative accuracy to the fully oracle results for the top-10 sentences.
The fully automatic method (our DenseVidCap) can consistently achieve more than 82\% relative accuracy of the oracle results on both top-5 and top-10 settings. This is very encouraging as the performance gap is not very large, especially considering that our model is trained with weakly annotated data. One important reason that causes the gap is that the evaluator network is not strong enough when compared with oracle re-ranking, which is a direction for further performance improvement.
\begin{table}[]
	\centering
    \small
	\resizebox{0.95\linewidth}{!}{
		\begin{tabular}{l|c|c|c|c}
			\hline
		\textbf{Model} & \textbf{METEOR} & \textbf{BLEU@4} & \textbf{ROUGE-L} & \textbf{CIDEr} \\ \hline
\multicolumn{4}{c}{Averaged Precision of Top-5 Sentences} \\  \hline
			\textbf{OSR + ORE}    &    29.3 (100)   &42.3 (100)   &64.1 (100) &  43.4 (100) \\
            \textbf{Dense + ORE}  &    28.0 (95.6)   &40.8 (96.5)   &62.8 (97.9) &  41.9 (96.5) \\
			\textbf{DenseVidCap}  &    26.5 (90.4)   &34.8 (82.3)   &57.7 (90.0) &  37.3 (85.9)  \\ \hline
\multicolumn{4}{c}{Averaged Precision of Top-10 Sentences} \\  \hline			
			\textbf{OSR + ORE}    &    27.9 (100)   &38.8 (100)   &61.4 (100) &  39.1 (100) \\
            \textbf{Dense + ORE}  &    26.4 (94.6)   &36.6 (94.3)   &59.5 (96.9) &  36.6 (93.6) \\
			\textbf{DenseVidCap}  &    26.1 (93.5)   &33.6 (86.6)   &57.1 (93.0) &  35.3 (90.3)  \\ \hline
		\end{tabular}
	}
    \vspace{-1 ex}
	\caption{Averaged precision of the top-5/10 sentences generated on the \textit{validation set} of MSR-VTT.
        \textbf{OSR} means oracle for sentence to region-sequence association, and \textbf{ORE} means oracle for sentence re-ranking.
        The values in the parenthesis indicate the relative percentage (\%) to the fully oracle results (OSR+ORE).
        }
	\label{top_res}
   \vspace{-0.15in}
\end{table}

\subsubsection{Diversity of Dense Captions}
The diversity of the generated captions is critical for dense video captioning.
We evaluate diversity from its opposite -- the similarity of the captions. A common solution is to determine the similarity between pairs of captions, or between one caption to a set of other captions.
Here we consider similarity from the apparent semantic relatedness of the sentences. We use the Latent semantic analysis (LSA)~\cite{deerwester1990indexing} which first generates sentence bag-of-words (BoW) representation, and then maps it to LSA space to represent a sentence.
This method has demonstrated its effectiveness in measuring document distance~\cite{kusner2015word}.
Based on the representation, we compute cosine similarity between two LSA vectors of sentences.
Finally, the diversity is calculated as:
\begin{equation}
\footnotesize
{D_{div}} = \frac{1}{n}\sum\limits_{\mathbf{s}^i, \mathbf{s}^j \in \mathbf{S}; ~i\neq j}(1 - \langle \mathbf{s}^i,\mathbf{s}^j \rangle ),
\end{equation}
where $\mathbf{S}$ is the sentence set with cardinality $n$, and $\langle \mathbf{s}^i,\mathbf{s}^j \rangle$ denotes the cosine similarity between $\mathbf{s}^i$ and $\mathbf{s}^j$.
\begin{figure}[]
	\centering
	\subfigure[]{\label{div:figa}
		\includegraphics[width=0.45\linewidth]{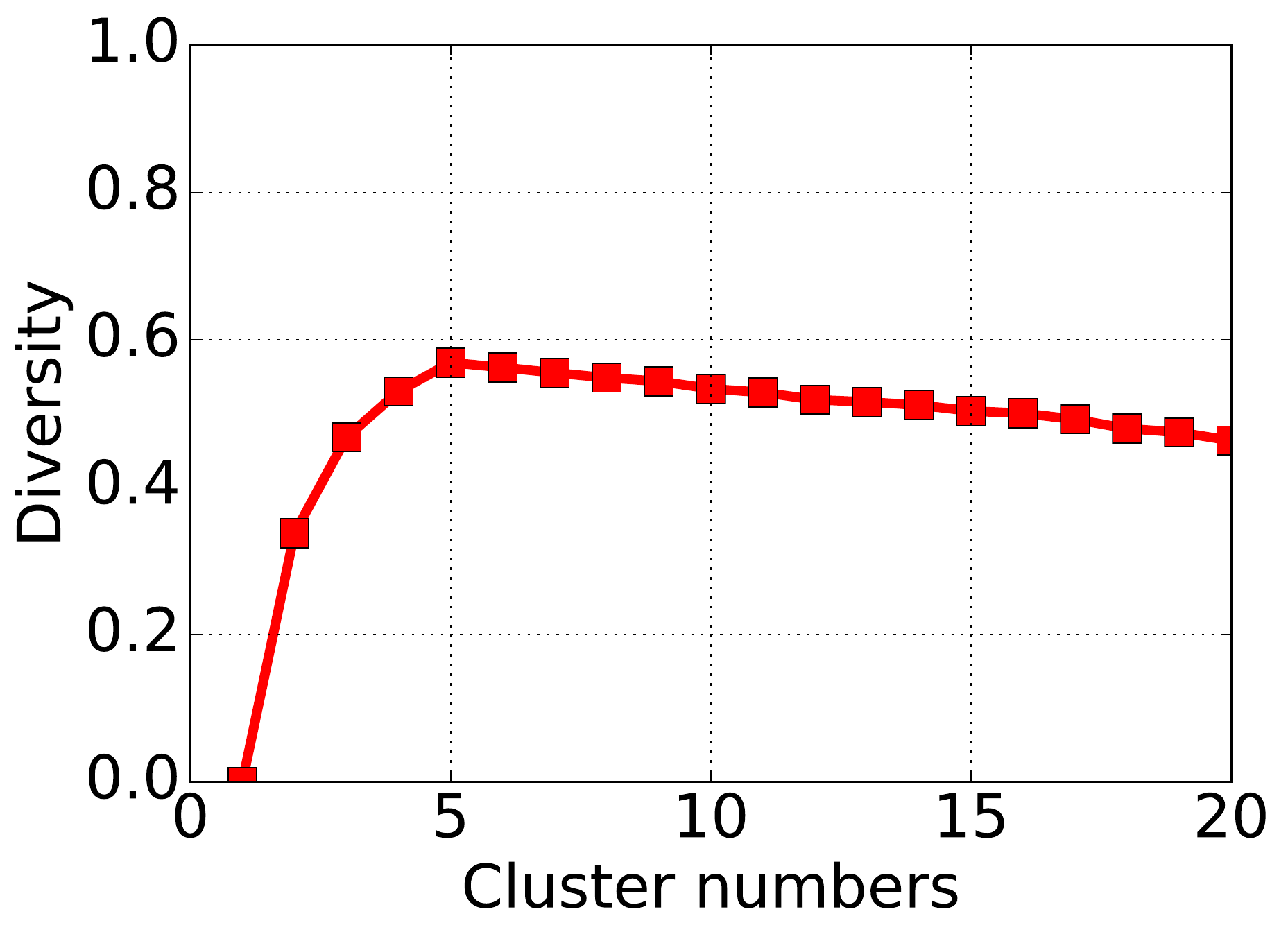}}
	\subfigure[]{\label{div:figb}
		\includegraphics[width=0.51\linewidth]{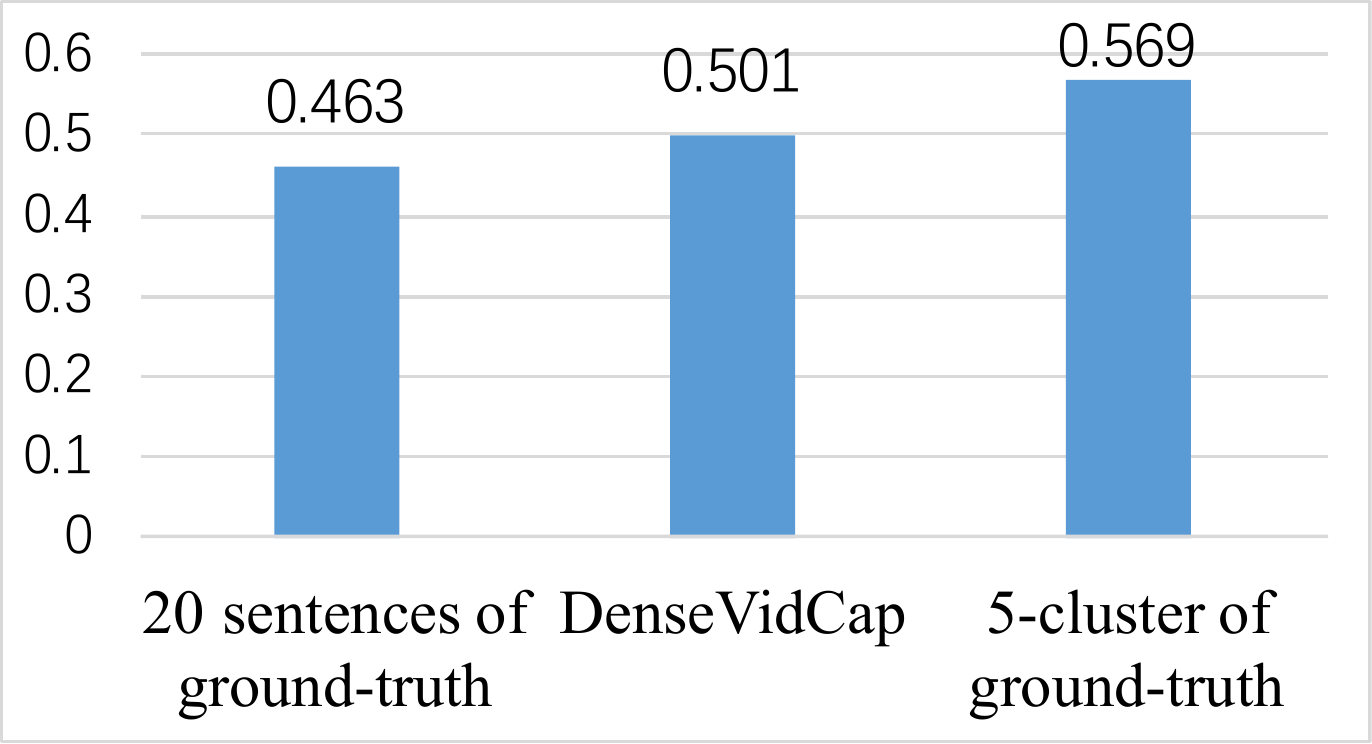}}
    \vspace{-2 ex}
	\caption{(a) Diversity score of clustered ground-truth captions under different cluster numbers; (b) Diversity score comparison of our automatic method (middle) and the ground-truth.}
	\label{diversity}
	\vspace{-0.15in}
\end{figure}

As aforementioned, we assume that the multiple video-level captions cover diversified aspects of the video content with some redundancy. The diversity metric can be applied in two aspects: evaluating the diversity degree of (1) our dense captioning results and (2) the manually generated captions in the ground-truth.
Some of the manually annotated ground-truth sentences on MSR-VTT are redundant. For instance, the captions ``a woman is surfing'' and ``a woman surfing in the ocean'' are more or less the same.
We remove the redundant captions by clustering on each video caption set with the LSA based representation.
Different clustering numbers can lead to different diversity scores. As shown in Figure~\ref{div:figa}, five clusters give the highest diversity score on this dataset.

We compare the diversity score of our automatic results with that of the ground-truth sentences in Figure~\ref{div:figb}.
As can be seen, our DenseVidCap achieves better diversity score (0.501) than that of the original 20 ground-truth sentences (0.463), but is slightly worse than that of the best of the clustered ground-truth sentences (0.569).
Please refer to Figure~\ref{demo} and the supplementary file for some qualitative dense video captioning results.
Both the diversity score and the qualitative results confirm that our proposed approach could produce diversified captioning output.

Through the comparison with the oracle results and the diversity evaluation in this subsection, we have demonstrated that our method can indeed produce good dense captions.

\section{Conclusion}
We have presented a weakly supervised dense video captioning approach, which is able to generate multiple diversified captions for a video clip with only video-level sentence annotations during the training procedure.
Experiments have demonstrated that our approach can produce multiple \textit{informative} and \textit{diversified} captions.
Our best single caption output outperforms the state-of-the-art methods on the MSR-VTT challenge with a significant margin.
Future work may consider leveraging the context among the dense captions to produce a consistent story for the input video clips.

\section*{Acknowledgements}
 Yu-Gang Jiang and Xiangyang Xue were supported in part by three NSFC projects (\#U1611461, \#61622204 and \#61572138) and a grant from STCSM, Shanghai, China (\#16JC1420401).

{\footnotesize
	\bibliographystyle{ieee}
	\bibliography{egbib}

\begin{thebibliography}{10}\itemsep=-1pt

\bibitem{anne2016deep}
L.~Anne~Hendricks, S.~Venugopalan, and et~al.
\newblock Deep compositional captioning: Describing novel object categories
  without paired training data.
\newblock In {\em CVPR}, 2016.

\bibitem{babenko2009visual}
B.~Babenko, M.-H. Yang, and S.~Belongie.
\newblock Visual tracking with online multiple instance learning.
\newblock In {\em CVPR}, 2009.

\bibitem{das2013thousand}
P.~Das, C.~Xu, and et~al.
\newblock A thousand frames in just a few words: Lingual description of videos
  through latent topics and sparse object stitching.
\newblock In {\em CVPR}, 2013.

\bibitem{deerwester1990indexing}
S.~Deerwester, S.~T. Dumais, and et~al.
\newblock Indexing by latent semantic analysis.
\newblock {\em Journal of the American society for information science},
  41(6):391, 1990.

\bibitem{donahue2015long}
J.~Donahue, L.~Anne, and et~al.
\newblock Long-term recurrent convolutional networks for visual recognition and
  description.
\newblock In {\em CVPR}, 2015.

\bibitem{dong2016early}
J.~Dong, X.~Li, and et~al.
\newblock Early embedding and late reranking for video captioning.
\newblock In {\em ACM Multimedia Grand Challenge}, 2016.

\bibitem{fang2015captions}
H.~Fang, S.~Gupta, and et~al.
\newblock From captions to visual concepts and back.
\newblock In {\em CVPR}, 2015.

\bibitem{girshick2015fast}
R.~Girshick.
\newblock Fast r-cnn.
\newblock In {\em ICCV}, 2015.

\bibitem{gygli2015video}
M.~Gygli, H.~Grabner, and L.~Van~Gool.
\newblock Video summarization by learning submodular mixtures of objectives.
\newblock In {\em CVPR}, 2015.

\bibitem{he2014spatial}
K.~He, X.~Zhang, S.~Ren, and J.~Sun.
\newblock Spatial pyramid pooling in deep convolutional networks for visual
  recognition.
\newblock In {\em ECCV}, 2014.

\bibitem{he2015deep}
K.~He, X.~Zhang, S.~Ren, and J.~Sun.
\newblock Deep residual learning for image recognition.
\newblock In {\em CVPR}, 2016.

\bibitem{heckerman2013tractable}
D.~Heckerman.
\newblock A tractable inference algorithm for diagnosing multiple diseases.
\newblock {\em arXiv:1304.1511}, 2013.

\bibitem{hochreiter1997long}
S.~Hochreiter and J.~Schmidhuber.
\newblock Long short-term memory.
\newblock {\em Neural computation}, 9(8), 1997.

\bibitem{jepson2003robust}
A.~D. Jepson, D.~J. Fleet, and T.~F. El-Maraghi.
\newblock Robust online appearance models for visual tracking.
\newblock {\em IEEE T PAMI}, 2003.

\bibitem{v2t_navigator}
Q.~Jin, J.~Chen, and et~al.
\newblock Describing videos using multi-modal fusion.
\newblock In {\em ACM Multimedia Grand Challenge}, 2016.

\bibitem{johnson2015densecap}
J.~Johnson, A.~Karpathy, and L.~Fei-Fei.
\newblock Densecap: Fully convolutional localization networks for dense
  captioning.
\newblock In {\em CVPR}, 2016.

\bibitem{karpathy2015deep}
A.~Karpathy and L.~Fei-Fei.
\newblock Deep visual-semantic alignments for generating image descriptions.
\newblock In {\em CVPR}, 2015.

\bibitem{khan2011human}
M.~U.~G. Khan, L.~Zhang, and Y.~Gotoh.
\newblock Human focused video description.
\newblock In {\em ICCV Workshops}, 2011.

\bibitem{kulkarni2013babytalk}
G.~Kulkarni, V.~Premraj, and et~al.
\newblock Babytalk: Understanding and generating simple image descriptions.
\newblock {\em IEEE T PAMI}, 2013.

\bibitem{kusner2015word}
M.~J. Kusner, Y.~Sun, and et~al.
\newblock From word embeddings to document distances.
\newblock In {\em ICML}, 2015.

\bibitem{lavie2014meteor}
M.~D.~A. Lavie.
\newblock Meteor universal: language specific translation evaluation for any
  target language.
\newblock {\em ACL}, 2014.

\bibitem{leskovec2007cost}
J.~Leskovec, A.~Krause, and et~al.
\newblock Cost-effective outbreak detection in networks.
\newblock In {\em ACM SIGKDD}, 2007.

\bibitem{lin2004rouge}
C.-Y. Lin.
\newblock Rouge: A package for automatic evaluation of summaries.
\newblock In {\em ACL-04 workshop on Text summarization branches out:}, 2004.

\bibitem{long2015fully}
J.~Long, E.~Shelhamer, and T.~Darrell.
\newblock Fully convolutional networks for semantic segmentation.
\newblock In {\em CVPR}, 2015.

\bibitem{lovasz1983submodular}
L.~Lov{\'a}sz.
\newblock Submodular functions and convexity.
\newblock In {\em Mathematical Programming The State of the Art}, pages
  235--257. Springer, 1983.

\bibitem{mao2014deep}
J.~Mao, W.~Xu, Y.~Yang, and et~al.
\newblock Deep captioning with multimodal recurrent neural networks (m-rnn).
\newblock {\em arXiv:1412.6632}, 2014.

\bibitem{maron1998framework}
O.~Maron and T.~Lozano-P{\'e}rez.
\newblock A framework for multiple-instance learning.
\newblock In {\em NIPS}, 1998.

\bibitem{nemhauser1978analysis}
G.~L. Nemhauser, L.~A. Wolsey, and M.~L. Fisher.
\newblock An analysis of approximations for maximizing submodular set
  functions—i.
\newblock {\em Mathematical Programming}, 14(1):265--294, 1978.

\bibitem{pan2015hierarchical}
P.~Pan, Z.~Xu, Y.~Yang, and et~al.
\newblock Hierarchical recurrent neural encoder for video representation with
  application to captioning.
\newblock In {\em CVPR}, 2016.

\bibitem{pan2015jointly}
Y.~Pan, T.~Mei, T.~Yao, and et~al.
\newblock Jointly modeling embedding and translation to bridge video and
  language.
\newblock In {\em CVPR}, 2016.

\bibitem{pancoast2014softening}
S.~Pancoast and M.~Akbacak.
\newblock Softening quantization in bag-of-audio-words.
\newblock In {\em IEEE ICASSP}, 2014.

\bibitem{papineni2002bleu}
K.~Papineni, S.~Roukos, T.~Ward, and et~al.
\newblock Bleu: a method for automatic evaluation of machine translation.
\newblock In {\em ACL}, 2002.

\bibitem{plummer2015flickr30k}
B.~A. Plummer, L.~Wang, C.~M. Cervantes, and et~al.
\newblock Flickr30k entities: Collecting region-to-phrase correspondences for
  richer image-to-sentence models.
\newblock In {\em ICCV}, 2015.

\bibitem{VideoLAB}
V.~Ramanishka, A.~Das, D.~H. Park, and et~al.
\newblock Multimodal video description.
\newblock In {\em ACM Multimedia Grand Challenge}, 2016.

\bibitem{yolo2016}
J.~Redmon, S.~Divvala, and et~al.
\newblock You only look once: Unified, real-time object detection.
\newblock In {\em CVPR}, 2016.

\bibitem{ren2015faster}
S.~Ren, K.~He, R.~Girshick, and J.~Sun.
\newblock Faster r-cnn: Towards real-time object detection with region proposal
  networks.
\newblock In {\em NIPS}, 2015.

\bibitem{rohrbach2014coherent}
A.~Rohrbach, M.~Rohrbach, and et~al.
\newblock Coherent multi-sentence video description with variable level of
  detail.
\newblock In {\em GCPR}, 2014.

\bibitem{rohrbach2015long}
A.~Rohrbach, M.~Rohrbach, and B.~Schiele.
\newblock The long-short story of movie description.
\newblock In {\em GCPR}, 2015.

\bibitem{russakovsky2015imagenet}
O.~Russakovsky, J.~Deng, H.~Su, J.~Krause, and et~al.
\newblock Imagenet large scale visual recognition challenge.
\newblock {\em IJCV}, 2015.

\bibitem{shetty2016frame}
R.~Shetty and J.~Laaksonen.
\newblock Frame-and segment-level features and candidate pool evaluation for
  video caption generation.
\newblock {\em arXiv:1608.04959}, 2016.

\bibitem{shin2016beyond}
A.~Shin, K.~Ohnishi, and et~al.
\newblock Beyond caption to narrative: Video captioning with multiple
  sentences.
\newblock {\em arXiv:1605.05440}, 2016.

\bibitem{simonyan2014very}
K.~Simonyan and A.~Zisserman.
\newblock Very deep convolutional networks for large-scale image recognition.
\newblock {\em arXiv:1409.1556}, 2014.

\bibitem{sutskever2014sequence}
I.~Sutskever, O.~Vinyals, and Q.~V. Le.
\newblock Sequence to sequence learning with neural networks.
\newblock In {\em NIPS}, 2014.

\bibitem{toutanova2003feature}
K.~Toutanova, D.~Klein, and et~al.
\newblock Feature-rich part-of-speech tagging with a cyclic dependency network.
\newblock In {\em NAACL}, 2003.

\bibitem{tran2015learning}
D.~Tran, L.~Bourdev, and et~al.
\newblock Learning spatiotemporal features with 3d convolutional networks.
\newblock In {\em ICCV}, 2015.

\bibitem{vedantam2015cider}
R.~Vedantam, C.~Lawrence~Zitnick, and D.~Parikh.
\newblock Cider: Consensus-based image description evaluation.
\newblock In {\em CVPR}, 2015.

\bibitem{venugopalan2016captioning}
S.~Venugopalan, L.~A. Hendricks, and et~al.
\newblock Captioning images with diverse objects.
\newblock {\em arXiv:1606.07770}, 2016.

\bibitem{venugopalan2015sequence}
S.~Venugopalan, M.~Rohrbach, and et~al.
\newblock Sequence to sequence-video to text.
\newblock In {\em ICCV}, 2015.

\bibitem{venugopalan2014translating}
S.~Venugopalan, H.~Xu, and et~al.
\newblock Translating videos to natural language using deep recurrent neural
  networks.
\newblock In {\em NAACL}, 2015.

\bibitem{vinyals2015show}
O.~Vinyals, A.~Toshev, and et~al.
\newblock Show and tell: A neural image caption generator.
\newblock In {\em CVPR}, 2015.

\bibitem{xu2016msr}
J.~Xu, T.~Mei, T.~Yao, and Y.~Rui.
\newblock Msr-vtt: A large video description dataset for bridging video and
  language.
\newblock In {\em CVPR}, 2016.

\bibitem{xu2015show}
K.~Xu, J.~Ba, and et~al.
\newblock Show, attend and tell: Neural image caption generation with visual
  attention.
\newblock {\em arXiv:1502.03044}, 2(3):5, 2015.

\bibitem{yao2015describing}
L.~Yao, A.~Torabi, K.~Cho, and et~al.
\newblock Describing videos by exploiting temporal structure.
\newblock In {\em ICCV}, 2015.

\bibitem{yu2015video}
H.~Yu, J.~Wang, and et~al.
\newblock Video paragraph captioning using hierarchical recurrent neural
  networks.
\newblock In {\em CVPR}, 2016.

\bibitem{zaremba2014recurrent}
W.~Zaremba, I.~Sutskever, and O.~Vinyals.
\newblock Recurrent neural network regularization.
\newblock {\em arXiv:1409.2329}, 2014.

\bibitem{zhang2005multiple}
C.~Zhang, J.~C. Platt, and P.~A. Viola.
\newblock Multiple instance boosting for object detection.
\newblock In {\em NIPS}, 2005.

\bibitem{zhou2006multi}
Z.-H. Zhou and M.-L. Zhang.
\newblock Multi-instance multi-label learning with application to scene
  classification.
\newblock In {\em NIPS}, 2006.

\end{thebibliography}
}
\newpage

{\noindent \Large \bf Supplementary Materials}
%\noindent \appendix{\large \textbf{Supplementary Materials}}\label{appendix}

\begin{appendices}
	
\section{Region-sequence Generation Algorithm}
Algorithm~\ref{inf} describes the region-sequence generation method, which is based on the CELF (Cost-Effective Lazy Forward selection) algorithm~\cite{leskovec2007cost}. 
In this algorithm, ${m}$ is the number of regions in a sequence, $UC$ and $CB$ are the abbreviation for uniform cost and cost benefit respectively.

\begin{algorithm}[h]
	\small
	\caption{ {{Region-sequence generation by submodular maximization with the CELF algorithm~\cite{leskovec2007cost}}}.}
	\label{inf}
	\begin{algorithmic}[1]
		\Function{$\mathbf{L_{AZY}F_{ORWARD}}$}{$\mathcal{S}_\mathbf{v}$, $x_\mathbf{v}$, $R$, $m$, $type$}
		\State $\mathcal{A} \leftarrow \emptyset;$
		\Comment{Start with the empty sequence}
		\For{each $r \in \mathcal{S}_\mathbf{v}$}
		$\mathcal{L}(\mathbf{w}; r) \leftarrow   \infty $
		\Comment{Init marginal gains}
		\EndFor
		\While{$\left| \mathcal{A} \right|  <  m$}
		\For{each $ r \in {\mathcal{S}_\mathbf{v}}\backslash \mathcal{A}$}
		${cur_s} \leftarrow $ \textbf{false};
		\EndFor
		\While{\textbf{true}}
		\Comment{Begin loop}
		\If{$type=UC$}
		\Comment{Uniform cost}
		\State ${r^*} \leftarrow \mathop {\arg \max }\limits_{r \in {\mathcal{S}_\mathbf{v}}\backslash \mathcal{A}} \ \mathcal{L}(\mathbf{w}; r);$
		\Comment{Max gain}
		\EndIf
		\If{$type=CB$}
		\Comment{Cost benefit}
		\State ${r^*} \leftarrow \mathop {\arg \max }\limits_{r \in {\mathcal{S}_\mathbf{v}}\backslash \mathcal{A}} \ \frac{{\mathcal{L}(\mathbf{w}; r)}}{{R(r)}};$
		\Comment{Max gain / cost}
		\EndIf
		\If{$cue_{r^*}$}
		$\mathcal{A} \leftarrow \mathcal{A} \cup \{r^*\};$ \textbf{break};
		%           \Comment{update set}
		\Else \
		\Comment{Update marginal gain}
		\State $\mathcal{L}(\mathbf{w}; r) \leftarrow R(\mathcal{A} \cup \{ r\} ) - R(\mathcal{A})$;
		\State $cur_{r^*} \leftarrow$ \textbf{true};
		\EndIf
		\EndWhile
		\EndWhile
		\State \Return $\mathcal{A}$;
		\Comment{Return region-sequence}
		\EndFunction
		\State
		\Function{$\mathbf{M_{AIN}}$}{$\mathcal{S}_\mathbf{v}$, $x_\mathbf{v}$, $R$, $m$}
		\State $\mathcal{A}_{UC} \leftarrow \mathbf{L_{AZY}F_{ORWARD}}$($\mathcal{S}_\mathbf{v}$, $x_\mathbf{v}$, $R$, $m$, $UC$)
		\State $\mathcal{A}_{CB} \leftarrow \mathbf{L_{AZY}F_{ORWARD}}$($\mathcal{S}_\mathbf{v}$, $x_\mathbf{v}$, $R$, $m$, $CB$)
		\State \Return $\arg \max \{ R({\mathcal{A}_{UC}}),\;R({\mathcal{A}_{CB}})\}$
		\EndFunction
	\end{algorithmic}
\end{algorithm}

\section{Response Maps}
\begin{figure}[h]
		\centering
		\includegraphics[width=0.95\linewidth]{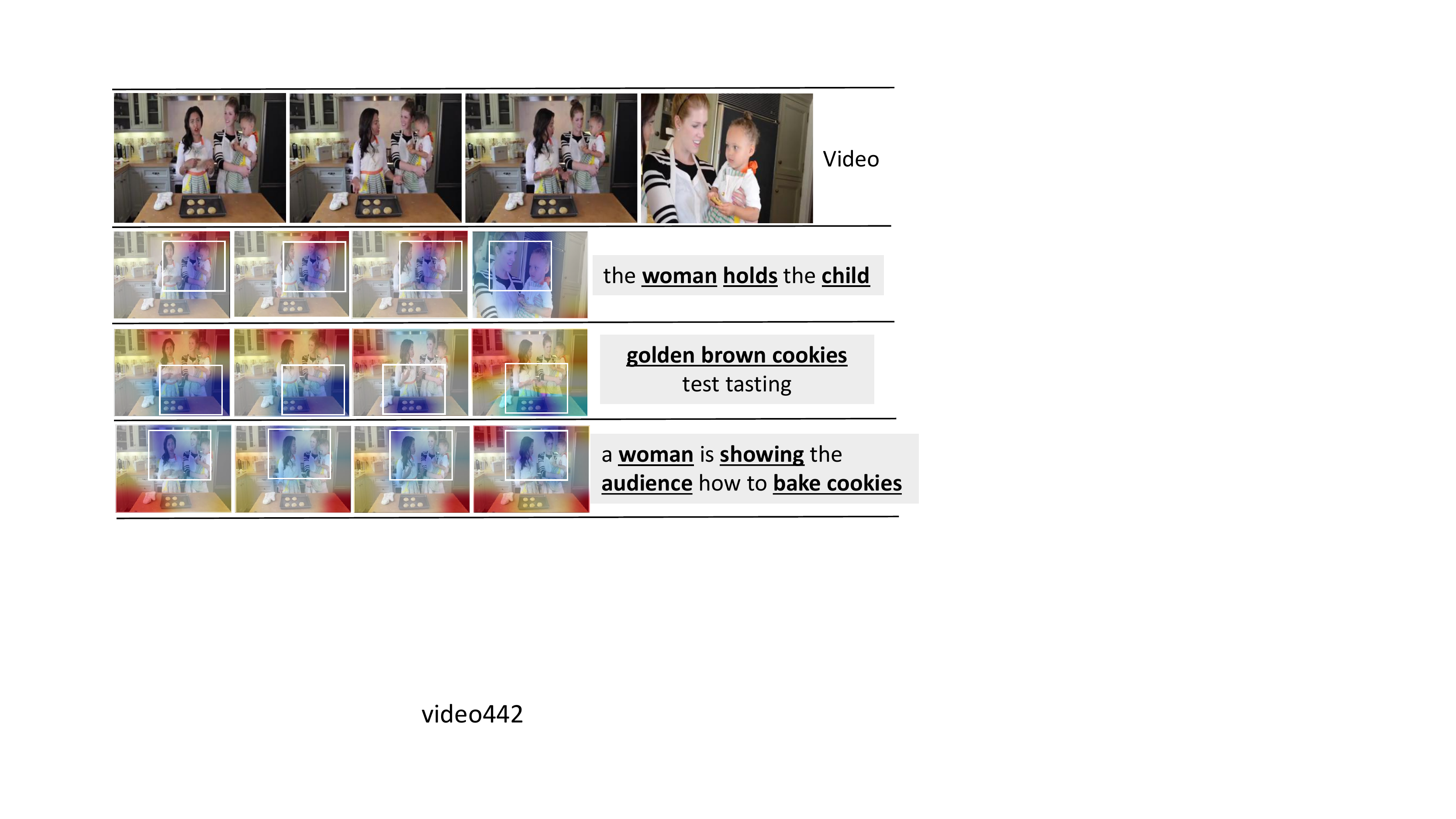}
		\vspace{-0.3 ex}
		\caption{
			Visualization of learned response maps from the last CNN layer (left), and the corresponding natural sentences (right).
			The blue areas in the response maps are of high attention, and the region-sequences are highlighted in white bounding-boxes.
		}
		\label{heatmap} 
		\vspace{-0.15in}
	\end{figure}

Figure~\ref{heatmap} shows some examples of response maps (heatmaps) generated by the Lexical-FCN model. 
We first associate the response maps to the words in the sentences based on the computed probabilities, and then visualize the best match.
	
\section{Sentence Re-ranking Module}
\begin{figure*}[t]
	\centering
	\includegraphics[width=0.9\linewidth]{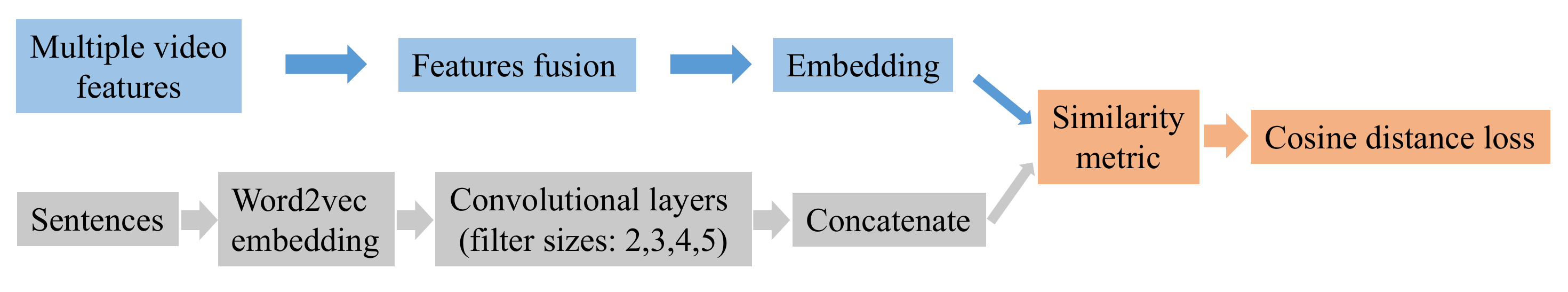}
	\vspace{-0.5 ex}
	\caption{Illustration of the sentence re-ranker module.}
	\label{re-ranker} 
	\vspace{-0.1in}
\end{figure*}

Figure~\ref{re-ranker} shows the diagram of our sentence re-ranking module, which re-rank multiple predicted sentences from dense video captioning. 
This module is similar to ~\cite{shetty2016frame}, which learns the cosine similarity between video features and sentence features with a neural network evaluator. 

\section{More Result Examples}	
\begin{figure*}[ht]
	\centering
	\includegraphics[width=0.95\linewidth]{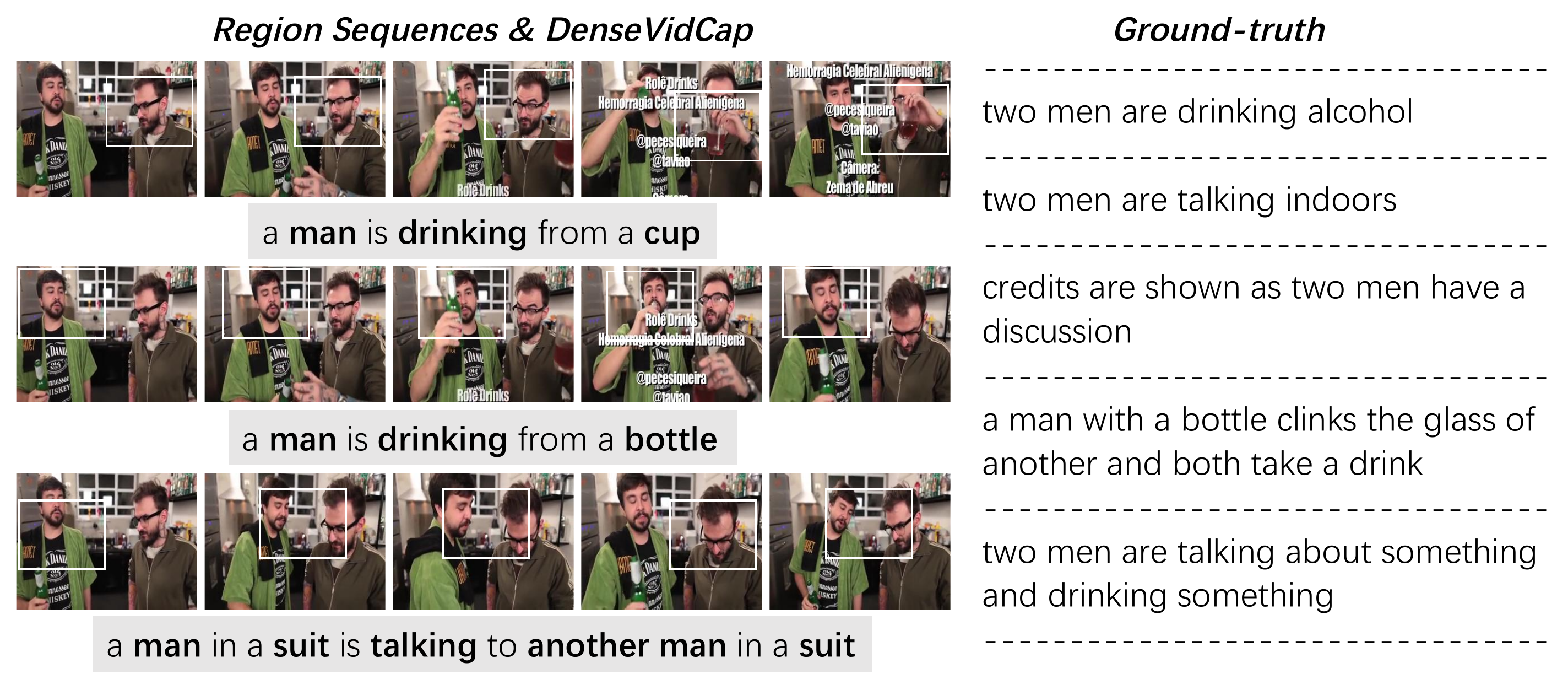}
	%\vspace{-0.5 ex}
	\caption{Left: Examples of dense sentences produced by our \textbf{{\em DenseVidCap}} method and corresponding \textbf{{\em region sequences}}; Right: Ground-truth (video6974).}
	\label{f0} 
	%\vspace{0.1in}
\end{figure*}

\begin{figure*}[ht]
	\centering
	\includegraphics[width=0.95\linewidth]{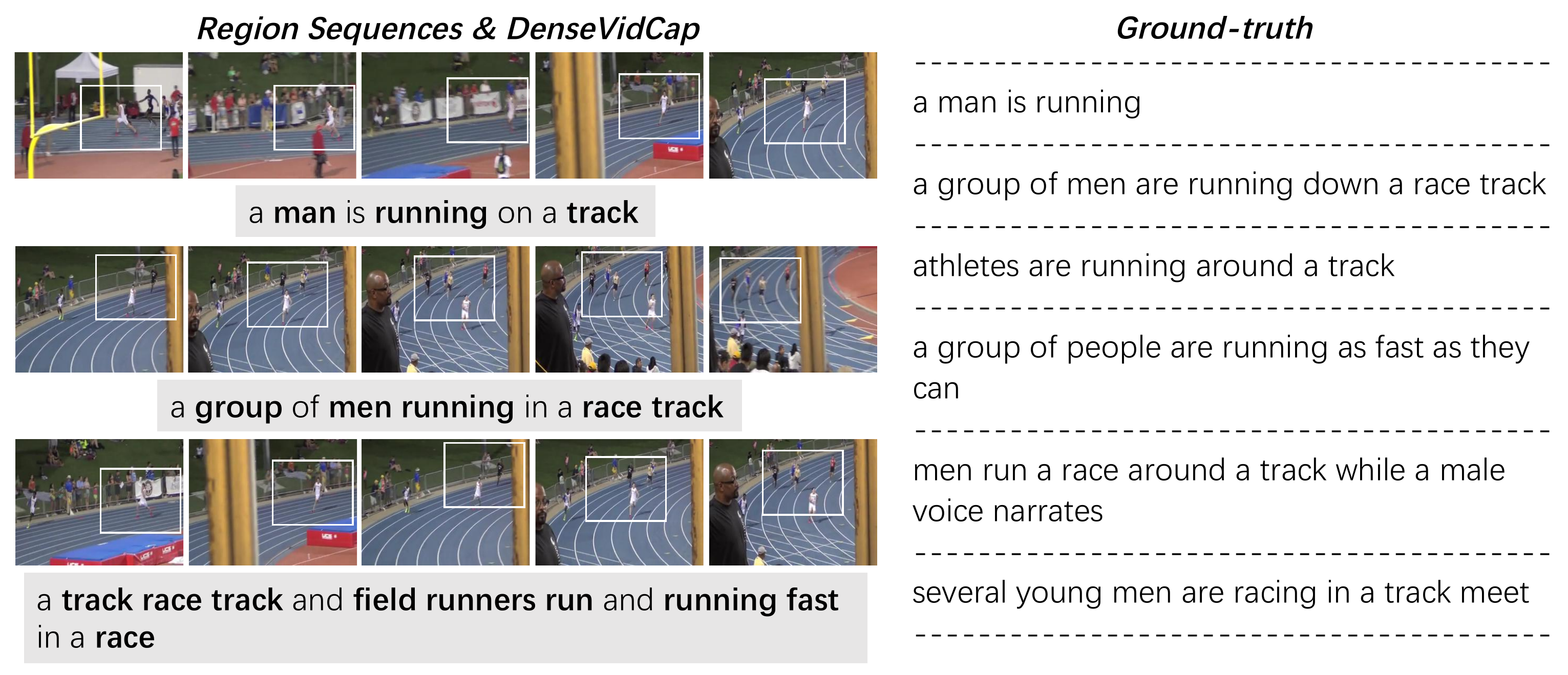}
	%\vspace{-0.5 ex}
	\caption{Left: Examples of dense sentences produced by our \textbf{{\em DenseVidCap}} method and corresponding \textbf{{\em region sequences}}; Right: Ground-truth (video6967).}
	\label{f1} 
	%\vspace{0.1in}
\end{figure*}

\begin{figure*}[t]
	\centering
	\includegraphics[width=0.95\linewidth]{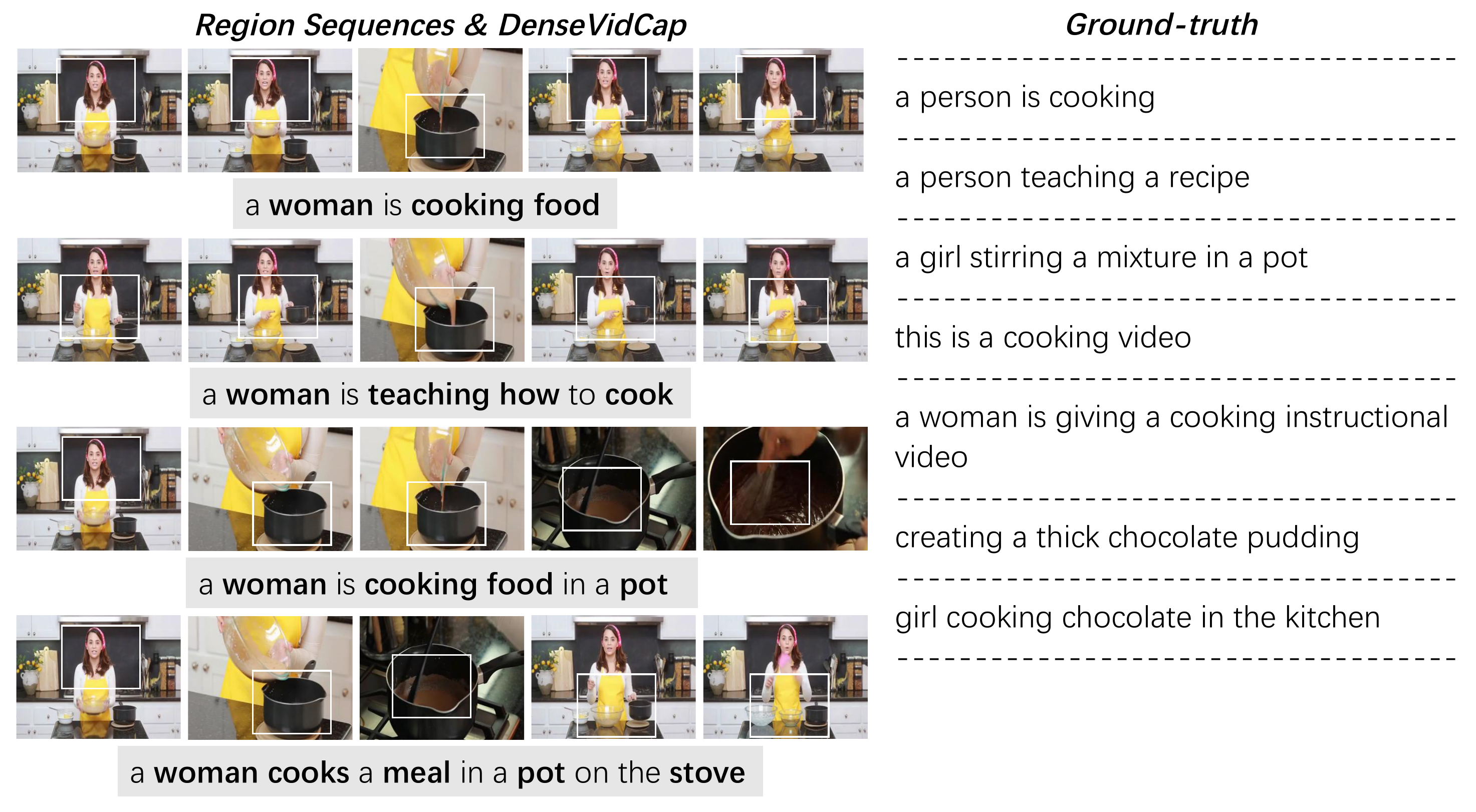}
	\vspace{-0.5 ex}
	\caption{Left: Examples of dense sentences produced by our \textbf{{\em DenseVidCap}} method and corresponding \textbf{{\em region sequences}}; Right: Ground-truth (video6911).}
	\label{f2} 
	%\vspace{0.1in}
\end{figure*}

\begin{figure*}[ht]
	\centering
	\includegraphics[width=0.95\linewidth]{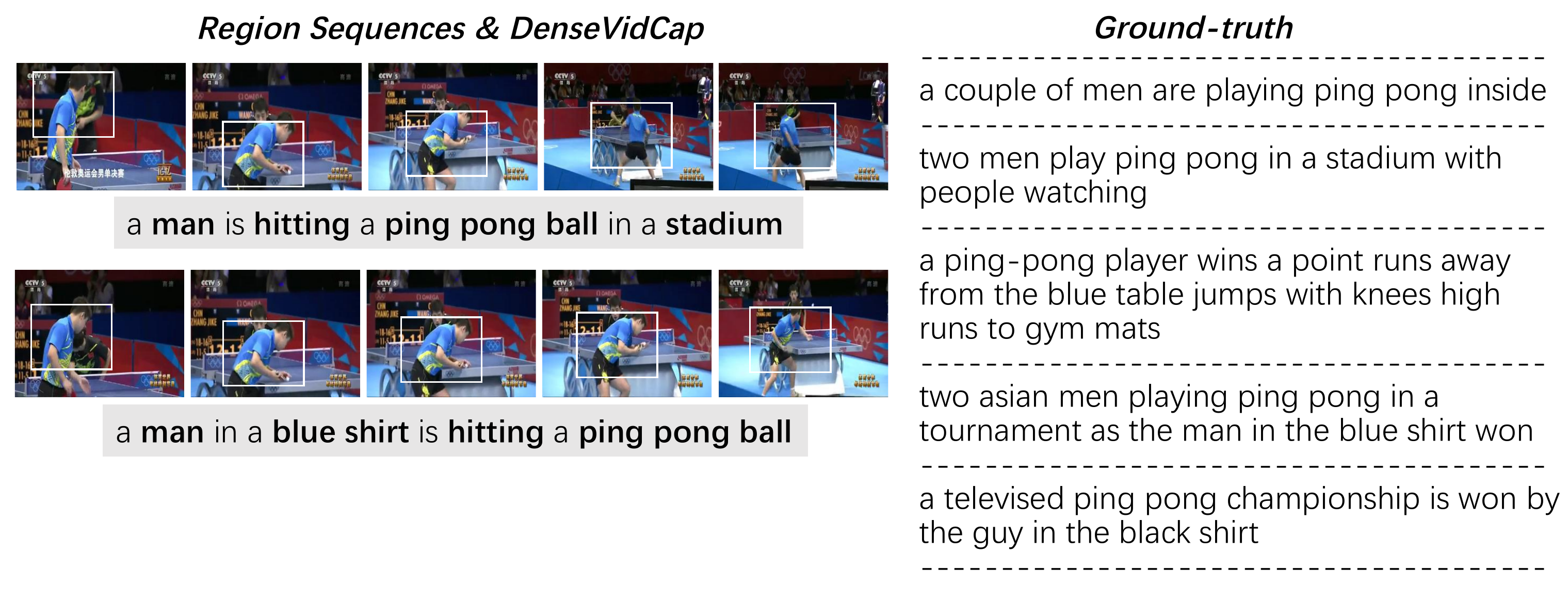}
	\vspace{-0.5 ex}
	\caption{Left: Examples of dense sentences produced by our \textbf{{\em DenseVidCap}} method and corresponding \textbf{{\em region sequences}}; Right: Ground-truth (video6973).}
	\label{f3} 
	%\vspace{0.1in}
\end{figure*}

More result examples of our DenseVideoCap system are provided in Figures~\ref{f0}, \ref{f1}, \ref{f2}, \ref{f3}.	

\end{appendices}

\end{document}